\pdfoutput=1

\documentclass[11pt]{article}

\usepackage{acl}

\usepackage{times}
\usepackage{latexsym}

\usepackage[T1]{fontenc}

\usepackage[utf8]{inputenc}

\usepackage{microtype}

\usepackage{inconsolata}

\usepackage{url}

\usepackage{booktabs}
\usepackage{multirow}
\usepackage{subcaption}

\usepackage{amssymb} 
\usepackage{amsmath}

\usepackage{tikz}
\usepackage{tikz-qtree}
\usetikzlibrary{arrows,decorations.pathmorphing,backgrounds,positioning,fit,petri,shapes.misc, arrows.meta,shapes.geometric,decorations.markings,calc,shadows.blur,decorations.pathreplacing,quotes}
\definecolor{myblue}{RGB}{6, 82, 221}
\definecolor{myorange}{RGB}{211, 84, 0}
\definecolor{lowblue}{RGB}{102,178,255}
\definecolor{justblue}{RGB}{84, 160, 255}
\definecolor{mypurple}{RGB}{108, 92, 231}
\definecolor{mygray}{RGB}{158, 158, 158}
\definecolor{lowpurple}{RGB}{204,153,255}
\definecolor{lowwhite}{RGB}{255,255,255}
\definecolor{verylowpurple}{RGB}{255,102,102}
\definecolor{embcolor}{RGB}{255,255,255}
\definecolor{myred}{RGB}{235, 47, 6} 
\definecolor{mygreen}{RGB}{162, 217, 206} 
\definecolor{fontgrey}{RGB}{44, 62, 80}
\definecolor{lowpurple}{RGB}{210, 180, 222}
\definecolor{mypumpkin}{RGB}{229, 152, 102}
\definecolor{lowgreen}{RGB}{171, 235, 198}
\definecolor{lowgreen2}{RGB}{186, 220, 88}
\definecolor{lowred}{RGB}{245, 183, 177}
\definecolor{lowyellow}{RGB}{241, 196, 15}
\definecolor{mypink}{RGB}{255, 118, 117}
\definecolor{bluemartina}{RGB}{18, 203, 196}
\definecolor{puffin}{RGB}{250, 152, 58}
\definecolor{grass}{RGB}{25, 197, 50}
\definecolor{grass2}{RGB}{0, 148, 60}
\definecolor{cnngray}{RGB}{116, 125, 140}
\definecolor{yelloworange}{RGB}{252, 150, 11}
\usepackage{color, colortbl}
\definecolor{Gray}{gray}{0.9}
\definecolor{lightgray}{gray}{0.95}
\definecolor{lightgray2}{gray}{0.9}

\newcommand{\squishlist}{
	\begin{list}{$\bullet$}
		{ \setlength{\itemsep}{0pt}
			\setlength{\parsep}{3pt}
			\setlength{\topsep}{3pt}
			\setlength{\partopsep}{0pt}
			\setlength{\leftmargin}{1.5em}
			\setlength{\labelwidth}{1em}
			\setlength{\labelsep}{0.5em} } }

	\newcounter{Lcount}
	\newcommand{\squishlisttwo}{
		\begin{list}{\arabic{Lcount}. }
			{ \usecounter{Lcount}
				\setlength{\itemsep}{0pt}
				\setlength{\parsep}{0pt}
				\setlength{\topsep}{0pt}
				\setlength{\partopsep}{0pt}
				\setlength{\leftmargin}{2em}
				\setlength{\labelwidth}{1.5em}
				\setlength{\labelsep}{0.5em} } }
		
		\newcommand{\squishend}{
	\end{list} }

\usepackage{multirow}
\usepackage{hhline}
\usepackage{tabularx}
\usepackage{ragged2e}
\newcolumntype{Y}{>{\RaggedRight\let\newline\\\arraybackslash\hspace{0pt}}X} 
\usepackage{xcolor}
\usepackage{pgfplots, pgfplotstable}
\pgfplotsset{compat=1.17}
\usepackage{pgfpages}
\usepackage{mathbbol}
\usepackage{arydshln}
\usepackage{graphicx}

\usepackage{amssymb}
\usepackage{amsfonts}

\usepackage{enumitem}
\usepackage{algorithm}
\usepackage{algpseudocode}
\usepackage{bbm}

\usepackage{pgf}
\usepackage{xspace}
\makeatletter
\DeclareRobustCommand\onedot{\futurelet\@let@token\@onedot}
\def\@onedot{\ifx\@let@token.\else.\null\fi\xspace}

\usepackage{lipsum}
\usepackage{multicol}
\usepackage{changepage}

\usepackage[export]{adjustbox}

\usepackage[utf8]{inputenc}
 
\usepackage{amsthm}
\usepackage{tablefootnote}
\usepackage{color,soul}
\setul{0.5ex}{0.3ex}
\setulcolor{blue}

\title{Exploring the Potential of Large Language Models in Computational Argumentation}

\author{
\textbf{
Guizhen Chen\thanks{Equal contribution. Guizhen Chen is under the Joint PhD Program between Alibaba and Nanyang Technological University.}
\textsuperscript{\rm 1,2}
\quad
Liying Cheng\footnotemark[1]\thanks{Liying Cheng is the corresponding author.}
\textsuperscript{\rm 1,3}
}
\textbf{
Luu Anh Tuan\textsuperscript{\rm 2} 
\quad
Lidong Bing\textsuperscript{\rm 1,3}
} \\
\textsuperscript{\rm 1}DAMO Academy, Alibaba Group, Singapore~~
\textsuperscript{\rm 2}Nanyang Technological University, Singapore \\
\textsuperscript{\rm 3}Hupan Lab, 310023, Hangzhou, China \\
{\tt\{guizhen.chen, liying.cheng, l.bing\}@alibaba-inc.com} \\
{\tt\{guizhen001, anhtuan.luu\}@ntu.edu.sg}
}

\begin{document}
\maketitle

\begin{abstract}
Computational argumentation has become an essential tool in various domains, including law, public policy, and artificial intelligence. It is an emerging research field in natural language processing that attracts increasing attention. Research on computational argumentation mainly involves two types of tasks: argument mining and argument generation. As large language models (LLMs) have demonstrated impressive capabilities in understanding context and generating natural language, it is worthwhile to evaluate the performance of LLMs on diverse computational argumentation tasks. This work aims to embark on an assessment of LLMs, such as ChatGPT, Flan models, and LLaMA2 models, in both zero-shot and few-shot settings. We organize existing tasks into six main categories and standardize the format of fourteen openly available datasets. In addition, we present a new benchmark dataset on counter speech generation that aims to holistically evaluate the end-to-end performance of LLMs on argument mining and argument generation.
Extensive experiments show that LLMs exhibit commendable performance across most of the datasets, demonstrating their capabilities in the field of argumentation.
Our analysis offers valuable suggestions for evaluating computational argumentation and its integration with LLMs in future research endeavors. \footnote{Our data and code implementation are released at \href{https://github.com/DAMO-NLP-SG/LLM-argumentation}{https://github.com/DAMO-NLP-SG/LLM-argumentation}.}

\end{abstract}

\section{Introduction}
Argumentation is a powerful and indispensable tool in various domains such as legality \cite{mochales2011argumentation, grabmair2015introducing}, debating \cite{slonim2021autonomous, li2020exploring}, and education \cite{Stab2016ParsingAS}. It plays a vital role in facilitating understanding between individuals by providing insights into different perspectives and their underlying reasons. Additionally, argumentation serves as a means of communicating convincing opinions, enhancing the acceptability of positions among readers.
As computational argumentation becomes a growing research field in natural language processing (NLP) \cite{Dietz2021ComputationalA, Habernal2016ArgumentationMI, Atkinson2017TowardsAA, Wachsmuth2017ComputationalAQ, Holtermann2022FairAA, Barrow2021SyntopicalGF}, researchers have dedicated considerable efforts to two distinct directions \cite{chakrabarty2019ampersand, cheng2021argument, alshomary2021b, bilu2019argument}.
The first direction, argument mining, focuses on understanding unstructured texts and automatically extracting various argumentative elements \cite{Cabrio2018FiveYO, levy2014context, Rinott2015ShowMY, Cheng2022IAMAC}. The other direction is argument generation, which aims to generate argumentative texts based on external knowledge \cite{Hua2019ArgumentGW, Schiller2020AspectControlledNA} or summarize key argument points. \cite{Syed2021GeneratingIC, Roush2020DebateSumAL}.

Unlike classical structure prediction NLP tasks like named entity recognition that typically take a single sentence as the input and extract token-level information, computational argumentation tasks require discourse-level comprehension. This requirement makes it challenging and laborious to gather a large volume of labeled data for training, hindering the progress of research in this field.
Fortunately, recent studies have shown that large language models (LLMs) \cite{Brown2020LanguageMA, Chowdhery2022PaLMSL, Tay2022UL2UL, Touvron2023LLaMAOA} have demonstrated impressive performance on a wide variety of NLP tasks \cite{Zhong2023CanCU,Pan2023APE, wang2023chatgpt, Cheng2023IsGA, shen2023large} in both zero-shot and few-shot settings.
Given their strong capability in understanding long contexts and generating natural language, it is exciting yet still questionable how well LLMs can perform computational argumentation tasks without any supervised training.

In light of this, our objective is to investigate the performance of LLMs on diverse computational argumentation tasks. There are two main issues we aim to address in our study. 
Firstly, although there are existing surveys about argument mining \cite{Peldszus2013FromAD}, the systematic study of the broader definition of computational argumentation including argument mining and argument generation is under-explored.
To bridge this gap, we categorize current computational argumentation tasks into two primary classes, comprising six distinct categories. In addition, we establish a standardized format and evaluation metrics for fourteen openly available datasets.
Secondly, existing tasks and datasets either focus on argument mining or argument generation.
To take a holistic approach, we propose a new task that integrates both argument mining and generation. This task is designed to generate counter speeches in response to debate speeches, which typically advocate a particular stance. We name them counter speech and supporting speech respectively in the remainder of our paper.
This task requires the model to understand the argumentative structures in the supporting speech, meanwhile to generate the counter speech against the proposition.
To facilitate the study, we construct a new document-to-document counterargument generation benchmark based on a debate database \cite{Lavee2019TowardsER}.

To evaluate the performance of LLMs on computational argumentation tasks, we choose from both open-source and proprietary LLMs to conduct our main experiments, in zero-shot and few-shot settings.
Our results reveal that LLMs exhibit promising performance in both argument mining and argument generation tasks. 
While LLMs might fail to achieve exceptionally high scores on specific metrics such as R\textsc{ouge}, we hypothesize that the strict nature of these metrics could potentially underestimate the true potential of LLMs, which are inherently generative in nature.
Human evaluation shows that LLMs are able to comprehend the core meaning of arguments and convey them effectively, even if the exact wording might not match.
Collectively, these findings highlight the strengths of LLMs in grasping and effectively conveying the essence of arguments, showcasing their potential beyond what traditional metrics may suggest.

To summarize, our contributions include:
\squishlist
\item{We organize the existing computational argumentation tasks including argument mining and argument generation, and standardize the format of related datasets.}
\item{We introduce a new task targeted at evaluating both argument mining and argument generation capabilities as a whole.}
\item{To the best of our knowledge, we for the first time systematically evaluate the performance of multiple computational argumentation tasks using LLMs in zero-shot and few-shot settings.}
\item{Extensive experimental results and analysis demonstrate the potential of LLMs in the computational argumentation research field and also suggest limitations in existing evaluation.}
\squishend

\section{Background}

\paragraph{Computational Argumentation}
\label{sec:ca}

Argumentation research has a long history \cite{Walton2008ArgumentationS, Hinton2019LanguageAA}, aiming to persuade through logical propositions and achieve agreement among parties \cite{Eemeren2003AST}. Recently, computational argumentation has emerged as a significant field in NLP. The two main research directions are argument mining and argument generation, along with other directions such as persuasiveness of arguments \cite{Habernal2016WhatMA} and quality assessment of arguments \cite{Wachsmuth2017ComputationalAQ}.
Our work specifically focuses on argument mining and argument generation, where the detailed background can be found in Appendix \ref{sec:appen_background}.

\paragraph{Large Language Models}
Recently, LLMs such as ChatGPT \cite{Chatgpt} have demonstrated strong capabilities in various NLP tasks. 
A surge of research has emerged to analyze and evaluate their performance on different types of tasks \cite{Leiter2023chatgptAM, Liu2023SummaryOC, Yang2023HarnessingTP, Guo2023HowCI, Laskar2023ASS}, including
translation \cite{Jiao2023IsCA,liu2024translation},
reasoning \cite{Shakarian2023AnIE, Frieder2023MathematicalCO, Liu2023EvaluatingTL, pan2023fact},
question answering \cite{Tan2023EvaluationOC,pham2024chatgpt},
sentiment analysis \cite{Zhong2023CanCU, Deng2023LLMsTT, Wang2023IsCA, Zhang2023SentimentAI,nguyen2023improving},
text-to-SQL \cite{Li2023CanLA, Liu2023ACE}, 
dialogue understanding \cite{Pan2023APE, Fan2023UncoveringTP, hu2023unlocking},
relation extraction \cite{Yuan2023ZeroshotTR,nguyen2023spectral,zheng2023jointprop},
hate speech detection \cite{Das2023EvaluatingCP,hoang2024toxcl},
summarization \cite{nguyen2022improving,Yang2023ExploringTL, Wang2023CrossLingualSV, Luo2023chatgptAA, Zhang2023ExtractiveSV}, and trustworthiness \cite{zhao2023prompt,zhao2024defending} etc.
However, it still lacks a systematic and thorough evaluation of computational argumentation using LLMs. 
Therefore, our work aims to explore the field of computational argumentation using LLMs by covering multiple tasks.

\section{Tasks and Datasets}
\begin{figure}[t]
    \centering
    \includegraphics[width=\columnwidth]{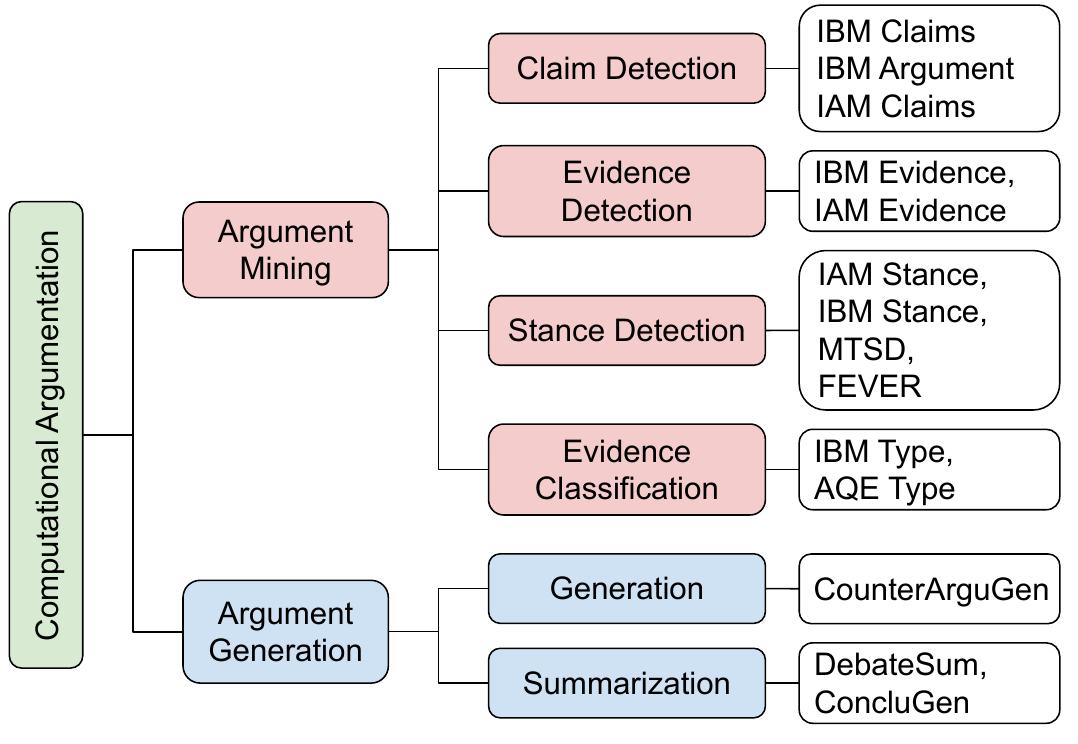}
    \caption{
    Explored tasks and datasets in this work.
    }
    \label{fig:tasks}
\end{figure}

In this work, we systematically review existing tasks and datasets of computational argumentation and organize them in Figure \ref{fig:tasks}.
To maintain a balance for different tasks and datasets, we restrict our assessment by randomly sampling 500 examples from each dataset.

\subsection{Argument Mining}

We focus on the detection of argumentative components and their relations, which are the fundamental tasks in argument mining. We include a range of datasets with varying levels of difficulty, from simple binary tasks such as claim detection and evidence detection to harder ones like evidence type classification and stance detection. While we do not cover joint tasks such as end-to-end argument mining \cite{eger-etal-2017-neural, bao-etal-2022-generative}, these tasks could be transformed into a sequence of subtasks of identifying argumentative components and relations, which could be handled by our evaluated tasks.

\paragraph{Claim Detection}
A claim is a statement or proposition that asserts something to be true or false.
In the context of argument mining, a claim is a key argument component that forms the basis of reasoning and debate.
In claim detection tasks, the goal is to automatically extract claims from articles related to a specific debating topic \cite{Levy2014ContextDC}.
We evaluate on datasets including IAM Claims \cite{Cheng2022IAMAC}, IBM Claims \cite{levy-etal-2018-towards}, and IBM Argument \cite{Shnarch2020UnsupervisedER}.

\paragraph{Evidence Detection}
Evidence is any information or data that supports or undermines a claim. 
In argument mining, evidence extraction involves automatically identifying and extracting relevant evidence from texts to substantiate claims \cite{Rinott2015ShowMY}.
Automating this process aids in comprehending and assessing arguments.
By pinpointing relevant evidence, researchers can gain valuable insights into the underlying beliefs and motivations behind an argument.
We evaluate evidence detection on the IBM Evidence dataset \cite{Shnarch2018WillIB} and the IAM Evidence dataset \cite{Cheng2022IAMAC}.

\paragraph{Stance Detection}
Stance represents a position towards a controversial topic, usually in the form of support and attack. Stance detection aims to determine whether a text supports, opposes, or remains neutral toward the topic. This task holds significance in domains such as politics \cite{Habernal2017TheAR}, fact-checking \cite{Thorne18Fever, Guo2021ASO}, and journalism \cite{Hanselowski2019ARA}, as it helps gauge public opinion and attitudes.
Automated stance detection enhances the understanding and analysis of arguments across various applications.
We use multiple datasets for evaluation, including FEVER \cite{Thorne18Fever}, IAM Stance \cite{Cheng2022IAMAC}, IBM Stance \cite{Levy2018TowardsAA}, and Multi-Target Stance Detection (MTSD) \cite{sobhani-etal-2017-dataset}.

\paragraph{Evidence Type Classification}
Evidence type refers to the different categories of evidence that can be used to support or undermine a claim \cite{Addawood2016WhatIY, Rinott2015ShowMY}. 
Examples of evidence types from previous works include statistics, expert opinions, facts, anecdotes, examples, etc.
Automatic evidence type classification aids in understanding the strengths and weaknesses of an argument, particularly in fields such as debate, law, and policy. 
We use two datasets for evaluation, including IBM Type \cite{Aharoni2014ABD} and AQE Type \cite{Guo2023AQEAQ}.

\subsection{Argument Generation}

We cover two main tasks: argument generation and argument summarization.

\paragraph{Generation}
Argument generation involves automatically generating arguments for or against a particular topic, to create persuasive and coherent arguments that can support or challenge a given position.
We adopt the CounterArguGen dataset \cite{alshomary2021b} for evaluation. There are two settings: generating a counter-argument given a claim with premises or generating based on a claim with weak premises.

\paragraph{Summarization}
The goal of argument summarization is to extract the main ideas and evidence supporting or challenging a particular claim or position and present them concisely and coherently.
We evaluate two datasets: ConcluGen \cite{Syed2021GeneratingIC} and DebateSum \cite{Roush2020DebateSumAL}, which aim to summarize or give a conclusion for arguments.
In the ConcluGen dataset, the corpus is augmented with three types of argumentative knowledge: topic, targets, and aspects. We study the effect of each argumentative knowledge and compare their respective performance with the base setting.
In the DebateSum dataset, there are two settings.
The abstractive summary generates a concise summary of the main points and arguments, while the extractive summary aims to extract relevant evidence from the passage to support the arguments.

\subsection{Counter Speech Generation}
Existing tasks in the field primarily center around either argument mining or argument generation. The former emphasizes language understanding, whereas the latter focuses on language generation. However, there is a lack of research comprehensively studying the overall argumentative capabilities of models. We contend that argument understanding and argument generation are two indispensable components of the broader computational argumentation landscape. 
Hence, a holistic perspective is necessary for evaluating the argumentative capabilities of models. Focusing solely on argument mining or argument generation provides only a partial understanding of their true potential.

In light of this, we propose a new task, counter speech generation, that aims to provide a more thorough evaluation of LLMs' argumentative capabilities. This task serves as a means to assess the model's capability to comprehend argumentative structures and generate counter-arguments accordingly.
In debates, a supporting speech serves as a form of discourse intended to construct a specific idea or stance. It aims to provide compelling arguments and evidence in favor of a particular viewpoint. Counter speech generation, therefore, involves the task of generating a responsive or opposing speech in reaction to the supporting speech.

To the best of our knowledge, this is the first document-to-document counterargument generation task that simultaneously assesses a model from multiple perspectives including claim detection, stance detection, and argument generation. Earlier works focus on mining and retrieval of counterarguments \cite{wachsmuth-etal-2018-retrieval, Bondarenko2020TouchFS, jo-etal-2021-knowledge-enhanced}, which does not involve argument generation. Some focus on generating an opposing argument for a given statement which are typically short, informal texts from online forums \cite{alshomary2021b, hua-wang-2018-neural, Hua2019ArgumentGW}. In contrast, ours consists of complete, formal speeches that are in the form of long argumentative texts, which potentially contain multiple arguments. Our task requires the model to first mine and analyze the main arguments in the original speech, then construct a complete and cohesive speech that addresses each key point. This expanded scope challenges the model to have a deeper understanding of argumentative structures from longer passages, while also requiring a heightened capacity to generate complete counter speeches.

To facilitate this study, we process a debate dataset \cite{Lavee2019TowardsER} by matching each supporting speech with the corresponding counter speech in a pool of debate scripts. We randomly sample 250 speech pairs for our zero-shot experiments. Given the constraint on limited annotated samples, we evaluate in a zero-shot setting only.
Appendix \ref{sec:sample} shows a data sample of this dataset.
\section{Experiments}

In this section, we discuss our choices of models, methods, and evaluation metrics.

\subsection{Models}
In our investigation, we examine the effectiveness of LLMs in directly performing inference on argument-related tasks without any fine-tuning. To accomplish this, we evaluate on open-source and proprietary models, including ChatGPT (\texttt{GPT-3.5-Turbo}) from OpenAI \cite{Chatgpt}, \texttt{Flan-T5-XL}, \texttt{Flan-T5-XXL} \cite{chung2022scaling} and \texttt{Flan-UL2} \cite{Tay2022UL2UL} from the Flan model family, as well as \texttt{Llama-2-7B}, \texttt{Llama-2-13B} from the LLaMA2 series \cite{LLaMA2}.

\subsection{Methods}
For tasks of a similar nature, we employ a consistent prompt format. 
More specifically, for argument mining tasks, we adhere to a standardized prompt template which consists of task definition and required output format. The task definition serves as a clear guideline for the LLMs to understand the task objective, while the required output format provides clarity on the expected output structure and restricts the generated response to a set of predefined labels to facilitate easier evaluation.

In contrast to argument mining tasks, output for argument generation tasks is more free-style and not constrained by any predetermined label space. The focus is on generating contextually relevant arguments. In order to tap into LLMs' linguistic knowledge and reasoning abilities, we adopt the prompts advised by ChatGPT.
The prompt templates are available in Appendix \ref{sec:prompts}.

\begin{figure}[t]
    \centering
    \includegraphics[width=1.0\columnwidth]{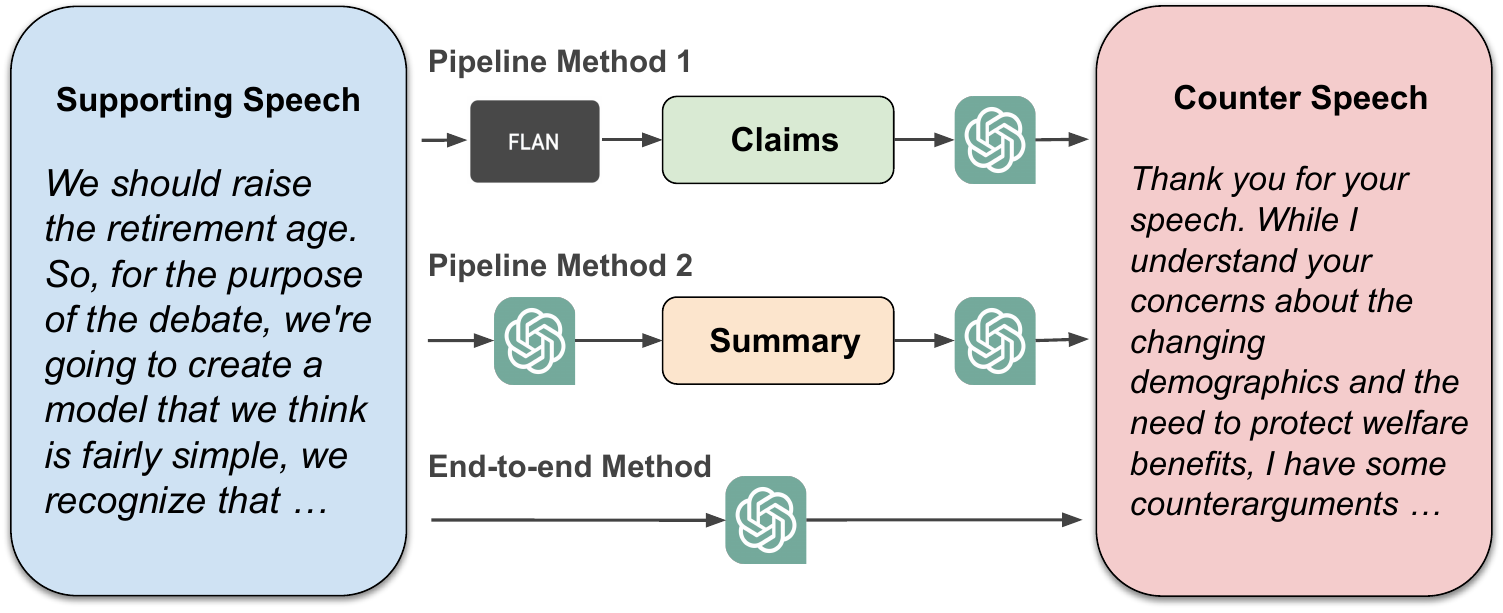}
    \caption{
    Three different approaches for our proposed task on counter speech generation.
    }
    \label{fig:flow}
\end{figure}

\begin{table*}[!t]
    \centering
    \resizebox{0.98\linewidth}{!}{
    \begin{tabular}{lccccccccccc}
    \toprule
    \multirow{2}{*}{\textbf{Model}} &
    \multicolumn{3}{c}{\textbf{Claim Detection}} &
    \multicolumn{2}{c}{\textbf{Evidence Detection}} &
    \multicolumn{4}{c}{\textbf{Stance Detection}} &
    \multicolumn{2}{c}{\textbf{Evidence Classification}} \\
    \cmidrule(lr){2-4}
    \cmidrule(lr){5-6}
    \cmidrule(lr){7-10}
    \cmidrule(lr){11-12} 
    & \textbf{\setulcolor{grass2}\ul{IBM Claims}} & \textbf{\setulcolor{grass2}\ul{IBM Argument}} & \textbf{\setulcolor{grass2}\ul{IAM Claims}} & 
    \textbf{\setulcolor{grass2}\ul{IBM Evidence}} & 
    \textbf{\setulcolor{grass2}\ul{IAM Evidence}} & \textbf{\setulcolor{grass2}\ul{IAM Stance}} & 
    \textbf{\setulcolor{grass2}\ul{IBM Stance}} & \textbf{\setulcolor{myred}\ul{MTSD}} & 
    \textbf{\setulcolor{myred}\ul{FEVER}} & 
    \textbf{\setulcolor{myred}\ul{IBM Type}} & 
    \textbf{\setulcolor{myred}\ul{AQE Type}} \\ 
    \midrule
    \multicolumn{11}{c}{$Acc.$} \\
    \midrule
    \rowcolor{lightgray2} 
    \texttt{Random} & 50.20 & 48.00 & 52.00 & 49.20 & 51.00 & 45.80 & 47.00 & 10.00 & 29.40 & 33.20 & 19.20 \\ 
    \texttt{GPT-3.5-Turbo} & 72.00 & 55.80 & 68.20 & 52.20 &
    45.00 & 59.00 & 33.80 & \textbf{41.00} & 33.40 & \textbf{73.40} & \textbf{58.20} \\ 
    \texttt{Flan-UL2 (20B)} & \textbf{74.80} & \textbf{63.60} & \textbf{83.80} & \textbf{64.80} & \textbf{71.40} & \textbf{65.00} & \textbf{58.20} & 15.40 & \textbf{35.40} & 68.60 & 21.60 \\ 
    \texttt{Llama-2-13B} & 36.00 & 44.20 & 44.80 & 25.40 &
    36.60 & 14.60 & 4.00 & 25.40 & 0.40 & 5.20 & 3.80 \\ 
    \midrule
    \multicolumn{11}{c}{$F_1$} \\
    \midrule
    \rowcolor{lightgray2} 
    \texttt{Random} & 55.53 & 52.01 & 64.28 & 49.62 & 58.05 & 51.76 & 50.05 & 12.59 & \textbf{33.93} & 33.51 & 24.42 \\
    \texttt{GPT-3.5-Turbo} & \textbf{72.19} & 56.16 & 76.35 & 50.44 & 51.48 & 58.99 & 36.26 & \textbf{42.27} & 20.33 & \textbf{72.39} & \textbf{59.95} \\
    \texttt{Flan-UL2 (20B)} & 71.80 & \textbf{62.06} & \textbf{86.80} & \textbf{64.45} & \textbf{75.70} & \textbf{63.71} & \textbf{59.70} & 13.38 & 28.06 & 67.34 & 15.68 \\
    \texttt{Llama-2-13B} & 40.61 & 41.73 & 56.84 & 21.99 & 46.05 & 18.30 & 6.28 & 12.51 & 0.77 & 8.22 & 4.59 \\
    \midrule
    \multicolumn{11}{c}{$p$-value} \\
    \midrule
    \texttt{GPT-3.5-Turbo} & 2.69e-11* & 1.61e-01 & 1.70e-08* & 6.71e-01 & 1.71e-03* & 1.27e-02* & 2.25e-07* & 0.00e+00* & 5.83e-01 & 0.00e+00* & 0.00e+00* \\
    \texttt{Flan-UL2 (20B)} & 5.86e-14* & 1.04e-04* & 0.00e+00* & 2.05e-04* & 2.67e-07* & 9.58e-06* & 1.35e-02* & 1.00e+00 & 2.19e-01 &	0.00e+00* & 1.88e-01 \\
    \texttt{Llama-2-13B} & 1.79e-09* & 1.29e-03* & 4.50e-01 & 9.47e-03* & 1.44e-11* & 0.00e+00* &	0.00e+00* &	3.44e-07* &	4.08e-01 & 0.00e+00* & 6.99e-15*\\
    \bottomrule
    \end{tabular}
    }
    \caption{Zero-shot performance on argument mining tasks. Datasets with binary class are underlined \setulcolor{grass2}\ul{green}. Datasets involving multi-class are underlined \setulcolor{myred}\ul{red}. Highest accuracy and F1 score for each task are in bold. $*$ indicates statistically significant results that are different from random observations with a significant level of $\alpha = 0.05$.}
    \label{tab:zero-shot-AM} 
\end{table*}

To tackle counter speech generation, we propose three different approaches\footnote{Note that other combinations of models could be used for each approach. Here we only employ the strongest model for each task/subtask, guided by the results in Section \ref{sec: results}.}, as shown in Figure \ref{fig:flow}. The first approach follows a pipeline method.
We first identify the main claims from the supporting speech by determining if each sentence is a claim towards the given topic.
We use \texttt{Flan-T5-XXL} due to its fast computation and strong capability in claim detection.
After identifying all claims, we generate counterarguments that attack each claim detected in the supporting speech. For this step, \texttt{GPT-3.5-Turbo} is employed due to its strong generative ability.

Another pipeline approach is by generating a summary of the supporting speech. Initially, the key arguments in the supporting speech are summarized into a condensed representation of the main points. Subsequently, a counter speech is crafted to challenge these key arguments. In both steps, we use \texttt{GPT-3.5-Turbo}, which is adept at handling long inputs and comprehending long contexts.

Unlike the two-step approaches, the third method is a one-step process where we directly prompt \texttt{GPT-3.5-Turbo} to respond to the supporting speech by challenging the main arguments. This approach serves as a means of gauging the model's ability to internally identify key arguments and generate a respective counter speech.

\subsection{Evaluation}
To evaluate argument mining tasks, we use both accuracy and F1 score as the metrics.

To assess argument generation and counter speech generation, we employ a wide range of automatic evaluation metrics, including R\textsc{ouge}-1, R\textsc{ouge}-2, R\textsc{ouge}-L  \cite{Lin2004ROUGEAP}, METEOR \cite{denkowski2011meteor} and BERTScore \cite{Zhang2019BERTScoreET}. The R\textsc{ouge} scores assess the quality based on the overlap with the reference arguments, while METEOR  also considers synonyms, paraphrases, and stemming. On the other hand, BERTScore takes into account the semantic context.  We also conduct human evaluation to complement the results from automatic evaluation.

\subsection{Previous SOTA}

To compare our results against existing state-of-the-art (SOTA), we either finetune pre-trained language models (PLMs) or
leverage available checkpoints to conduct inference on our sampled test set. Training details are reported in Appendix \ref{sec:training}.

\begin{figure*}[t!]
\captionsetup[subfigure]{justification=centering}
\begin{subfigure}[t]{0.25\linewidth}
\centering
\begin{tikzpicture}
\pgfplotsset{width=6cm,height=5cm,compat=1.8}
\begin{axis}[
    xtick={1,2,3,4,5},
    ymin=5, ymax=85,
    xticklabels = {0,1,3,5,10},
    xticklabel style = {font=\fontsize{6}{1}\selectfont},
    yticklabel style = {font=\fontsize{6}{1}\selectfont},
    legend style={font=\fontsize{7}{1}\selectfont},
	ylabel={\footnotesize F1 Score},
	xlabel={\footnotesize k-shot for simple tasks*},
	enlargelimits=0.1,
    legend style={at={(1.6, 1.05)},anchor=south,legend columns=6}, 
	every axis plot/.append style={thick},
	tick label style={/pgf/number format/fixed},
    every node near coord/.append style={font=\tiny}
]

 \addplot[lowyellow] [mark=triangle,mark size=2.7pt, thin] coordinates {
(1, 60.9548908815424)
(2, 61.828354605535594)
(3, 60.6560019629247)
(4, 60.63752864820766)
(5, 59.802809315538)
 };

 \addplot[yelloworange] [mark=triangle*,mark size=2.7pt, thin] coordinates {
(1, 67.29176994203723)
(2, 67.96242007813598)
(3, 68.77297687057367)
(4, 68.02829510683782)
(5, 67.02638328289503)
};

 \addplot[myorange] [mark=triangle*,mark size=2.7pt, thin] coordinates {
(1, 70.37130642365862)
(2, 71.92609889383215)
(3, 72.60991863740396)
(4, 73.3116611626043)
(5, 73.5301973570907)
};

\addplot[lowblue]  [mark=square*, thin]  coordinates {
(1, 57.6194181453319)
(2, 59.053521150207914)
(3, 63.33176862157253)
(4, 64.33985812092186)
(5, 65.55487205158357)
};

 \addplot[grass] [mark=otimes, thin] coordinates {
(1, 31.65621217285936)
(2, 39.18623312772721)
(3, 38.60934289574887)
(4, 37.29785452624595)
(5, 34.12893136050351)
};

 \addplot[grass2] [mark=otimes*,thin] coordinates {
(1, 31.67765717077039)
(2, 24.590690359406032)
(3, 36.36119284947038)
(4, 33.99964732578701)
(5, 40.670091946761495)
};

\addplot[black, dotted]
 coordinates {
(1, 82.3)
(2, 82.3)
(3, 82.3)
(4, 82.3)
(5, 82.3)
};

\addplot[gray, dotted] 
 coordinates {
(1, 71.5)
(2, 71.5)
(3, 71.5)
(4, 71.5)
(5, 71.5)
};

\legend{{\texttt{Flan-T5-XL}}, {\texttt{Flan-T5-XXL}}, {\texttt{Flan-UL2}}, {\texttt{GPT-3.5-Turbo}}, {\texttt{Llama-2-7B}}, {\texttt{Llama-2-13B}}, SOTA (full setting), SOTA (500 samples)}
\end{axis}
\end{tikzpicture}
\label{fig:few_shot_simple}
\end{subfigure}
~~~~~~~~~~~~~
\begin{subfigure}[t]{0.25\linewidth}
\centering
\begin{tikzpicture}
\pgfplotsset{width=6cm,height=5cm,compat=1.8}
\begin{axis}[
    xtick={1,2,3,4,5},
    ymin=5, ymax=85,
    xticklabels = {0,1,3,5,10},
    xticklabel style = {font=\fontsize{6}{1}\selectfont},
    yticklabel style = {font=\fontsize{6}{1}\selectfont},
    legend style={font=\fontsize{5}{1}\selectfont},
	xlabel={\footnotesize k-shot for hard tasks},
	enlargelimits=0.1,
	legend style={at={(0.5, 1.05)},anchor=south,legend columns=3}, 
	every axis plot/.append style={thick},
	tick label style={/pgf/number format/fixed},
    every node near coord/.append style={font=\tiny}
]

 \addplot[lowyellow] [mark=triangle,mark size=2.7pt, thin] coordinates {
(1, 32.600460451105256)
(2, 34.4248199884591)
(3, 36.44560747537067)
(4, 35.788410420347546)
(5, 36.60601540754113)
 };

 \addplot[yelloworange] [mark=triangle*,mark size=2.7pt, thin] coordinates {
(1, 48.02884074799471)
(2, 50.299169616101004)
(3, 51.9473617128605)
(4, 51.54424595969631)
(5, 52.96331971203115)
};

 \addplot[myorange] [mark=triangle*,mark size=2.7pt, thin] coordinates {
(1, 31.11288058485759)
(2, 35.70715958454148)
(3, 36.59717733613021)
(4, 37.66334566444955)
(5, 38.20447715633964)
};

\addplot[lowblue]  [mark=square*, thin]  coordinates {
(1, 48.73368378338769)
(2, 46.36070736233297)
(3, 51.87529289264386)
(4, 54.673297207099445)
(5, 55.39905201827502)
};

 \addplot[grass] [mark=otimes,thin] coordinates {
(1, 7.573723846923033)
(2, 7.6131767548041465)
(3, 5.565895261651032)
(4, 6.985118334923604)
(5, 11.872596595107948)
};

 \addplot[grass2] [mark=otimes*,thin] coordinates {
(1, 6.524658235115892)
(2, 21.502574257879502)
(3, 19.099534335261477)
(4, 28.670286710804398)
(5, 30.26460674815671)
};

\addplot[black, dotted]
 coordinates {
(1, 61)
(2, 61)
(3, 61)
(4, 61)
(5, 61)
};

\addplot[gray, dotted]
 coordinates {
(1, 53)
(2, 53)
(3, 53)
(4, 53)
(5, 53)
};

\end{axis}
\end{tikzpicture}
\label{fig:few_shot_hard}
\end{subfigure}
~~~~~~~~
\begin{subfigure}[t]{0.25\linewidth}
\centering
\begin{tikzpicture}
\pgfplotsset{width=6cm,height=5cm,compat=1.8}
\begin{axis}[
    xtick={1,2,3,4,5},
    ymin=5, ymax=85,
    xticklabels = {0,1,3,5,10},
    xticklabel style = {font=\fontsize{6}{1}\selectfont},
    yticklabel style = {font=\fontsize{6}{1}\selectfont},
    legend style={font=\fontsize{5}{1}\selectfont},
	xlabel={\footnotesize k-shot for all tasks*},
	enlargelimits=0.1,
	legend style={at={(0.5, 0.9)},anchor=south,legend columns=2}, 
	every axis plot/.append style={thick},
	tick label style={/pgf/number format/fixed},
    every node near coord/.append style={font=\tiny}
]

\addplot[lowblue]  [mark=square*, thin]  coordinates {
(1, 54.06512440055422)
(2, 53.97639563505794)
(3, 58.74917833000107)
(4, 60.4732337553929)
(5, 61.49254403826014)
};

 \addplot[lowyellow] [mark=triangle,mark size=2.7pt, thin] coordinates {
(1, 49.61311870936754)
(2, 50.866940758705006)
(3, 50.97184416790309)
(4, 50.697881357063615)
(5, 50.52409175233925)
 };

 \addplot[yelloworange] [mark=triangle*,mark size=2.7pt, thin] coordinates {
(1, 59.58659826442023)
(2, 60.89711989332199)
(3, 62.0427308074884)
(4, 61.43467544798121)
(5, 61.40115785454948)
};

 \addplot[myorange] [mark=triangle*,mark size=2.7pt, thin] coordinates {
(1, 54.667936088138205)
(2, 57.438523170115886)
(3, 58.20482211689446)
(4, 59.05233496334243)
(5, 59.399909276790275)
};

 \addplot[grass] [mark=otimes,thin] coordinates {
(1, 22.023216842484832)
(2, 26.55701057855799)
(3, 25.391963842109732)
(4, 25.799920108847818)
(5, 25.226397454345285)
};

 \addplot[grass2] [mark=otimes*,thin] coordinates {
(1, 21.616457596508592)
(2, 23.35544391879542)
(3, 29.456529443786817)
(4, 31.86790307979396)
(5, 36.50789786731958)
};

\addplot[black, dotted]
 coordinates {
(1, 72)
(2, 72)
(3, 72)
(4, 72)
(5, 72)
};

\addplot[gray, dotted]
 coordinates {
(1, 64)
(2, 64)
(3, 64)
(4, 64)
(5, 64)
};

\end{axis}
\end{tikzpicture}
\label{fig:few_shot_all}
\end{subfigure}

\vspace{-6mm}
\caption{Few-shot performance comparison on argument mining tasks. Results of the previous SOTA (using full setting and 500 samples) are also shown for easy comparison. *: Note that we exclude \textit{IBM Argument} because the train set is smaller than 500.}
\label{fig: task_difficulty}
\end{figure*}
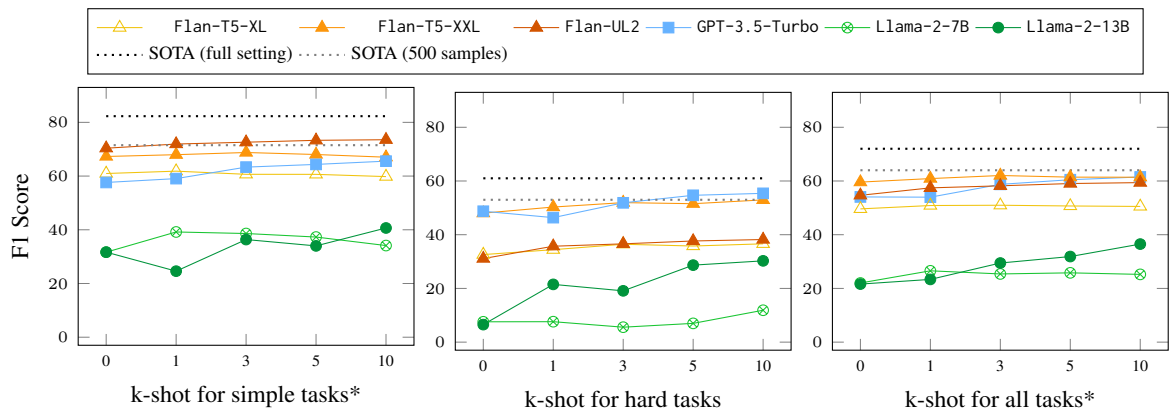

\section{Results and Discussion}
\label{sec: results}
In this section, we discuss the main results and provide insights into the performance of various LLMs on argument mining, argument generation, and counter speech generation.

\subsection{Results on Argument Mining}
Table \ref{tab:zero-shot-AM} shows the zero-shot performance of three representative models, \texttt{GPT-3.5-Turbo}, \texttt{Flan-UL2} and \texttt{Llama-2-13B}, across 11 argument mining datasets. Statistical tests\footnote{We use McNemar’s test \cite{Mcnemar1947NoteOT} following the guidelines by \citet{dror-etal-2018-hitchhikers}.} are conducted to show if the LLM's predictions are significantly different from the random observations.
Results of other models including \texttt{Flan-T5-XL}, \texttt{Flan-T5-XXL} and \texttt{Llama-2-7B} are available in Appendix \ref{sec:appen_am_few}.

To analyze, we categorize tasks into \setulcolor{grass2}\ul{simple} and \setulcolor{myred}\ul{hard} tasks based on the number of classes involved. 
Binary classification tasks, including claim detection, evidence detection, IAM stance detection, and IBM stance detection, are classified as \setulcolor{grass2}\ul{simple} tasks. 
Tasks with more than two labels, including evidence classification, FEVER stance detection, and MTSD stance detection, are classified as \setulcolor{myred}\ul{hard} tasks.

Overall, both \texttt{GPT-3.5-Turbo} and \texttt{Flan-UL2} perform decently in the zero-shot setting, surpassing the random baseline with most results being statistically different.
However, \texttt{Llama-2-13B} falls short of the random baseline in the majority of the tasks, notably in more challenging tasks like FEVER stance detection and evidence classification. This highlights its limitations in capturing nuanced stances and comprehending evidence types within the zero-shot context, which relies on sufficient prior knowledge in the model.

Comparing \texttt{GPT-3.5-Turbo} and \texttt{Flan-UL2}, \texttt{Flan-UL2} consistently demonstrates higher proficiency in tasks like claim detection, evidence detection, and certain stance detection tasks that are mostly binary classification tasks. However, its performance diminishes when encountered with tasks that involve more than two classes, such as MTSD stance detection and AQE evidence classification. In contrast, \texttt{GPT-3.5-Turbo} generally demonstrates superior performance in these multi-class scenarios.

\begin{table*}[!t]
    \centering
    \resizebox{0.98\linewidth}{!}{
    \begin{tabular}{llllclllll}
    \toprule
    \textbf{Task} & \textbf{Dataset} & \textbf{Setting} & \textbf{Method} & \textbf{k-shot} & \textbf{BERTScore} & \textbf{R\textsc{ouge}1} & \textbf{R\textsc{ouge}2} & \textbf{R\textsc{ouge}L} & \textbf{METEOR} \\
    \midrule
    \multirow{4}{*}{Generation}  & 
    \multirow{4}{*}{CounterArguGen} & 
    \multirow{2}{*}{Premises} & 
    \cellcolor{lightgray2}\citet{alshomary2021b} & 
    \cellcolor{lightgray2}- & 
    \cellcolor{lightgray2}$82.60$	& 
    \cellcolor{lightgray2}$17.76$	& 
    \cellcolor{lightgray2}$1.36$	&  
    \cellcolor{lightgray2}$10.66$	& 
    \cellcolor{lightgray2}$14.85$ \\
    &&& \texttt{GPT-3.5-Turbo} & k=0 & $83.50$&	$18.36$&	$1.58$&	$11.07$&	$17.60$  \\
    \cmidrule{3-10}
     && \multirow{2}{*}{Weak Premises}& \cellcolor{lightgray2}\citet{alshomary2021b} & 
     \cellcolor{lightgray2}- &
     \cellcolor{lightgray2}$82.53$ & 
     \cellcolor{lightgray2}$17.34$ & 
     \cellcolor{lightgray2}$1.12$ & 
     \cellcolor{lightgray2}$10.33$ & 
     \cellcolor{lightgray2}$14.65$ \\
     &&& \texttt{GPT-3.5-Turbo} & k=0 & $84.06$&	$19.75$&	$2.03$&	$11.95$&	$17.63$  \\
    \midrule
    \multirow{30}{*}{Summarization} &
    \multirow{20}{*}{ConcluGen} &
    \multirow{5}{*}{Base} 
    & \cellcolor{lightgray2}\citet{Syed2021GeneratingIC}& \cellcolor{lightgray2}- &
    \cellcolor{lightgray2}$84.78$ & 
    \cellcolor{lightgray2}$8.16$ & 
    \cellcolor{lightgray2}$0.47$ & 
    \cellcolor{lightgray2}$7.15$ & 
    \cellcolor{lightgray2}$6.02$ \\
    &&&\texttt{GPT-3.5-Turbo} &k=0 & $85.53$&	$13.99$&	$3.20$&	$10.78$&	$21.28$  \\
    &&&\texttt{GPT-3.5-Turbo} & k=1 
    & $86.51_{0.15}$  
    & $16.80_{0.33}$  
    &  $3.86_{0.18}$  
    & $12.96_{0.38}$  
    & $20.34_{0.24}$ \\
    &&&\texttt{GPT-3.5-Turbo} & k=3 
    & $86.95_{0.20}$ 
    & $18.54_{0.91}$ 
    &  $4.66_{0.52}$  
    & $14.53_{0.78}$ 
    & $21.27_{0.59}$  \\
    &&&\texttt{GPT-3.5-Turbo} & k=5 
    & $87.19_{0.27}$  
    & $19.39_{0.78}$ 
    &  $5.09_{0.47}$  
    & $15.21_{0.85}$ 
    & $21.50_{0.11}$ \\
    \cmidrule{3-10}
    &&\multirow{5}{*}{Aspects} & \cellcolor{lightgray2}\citet{Syed2021GeneratingIC}& 
    \cellcolor{lightgray2}- &
    \cellcolor{lightgray2}$89.32$ & 
    \cellcolor{lightgray2}$31.47$ & 
    \cellcolor{lightgray2}$16.90$ & 
    \cellcolor{lightgray2}$28.94$ & 
    \cellcolor{lightgray2}$27.61$\\
    &&&\texttt{GPT-3.5-Turbo} &k=0 & $85.47$&	$13.79$&	$3.25$&	$10.43$&	$21.70$  \\
    &&& \texttt{GPT-3.5-Turbo} &k=1
    & $86.16_{0.25}$
    & $16.41_{0.92}$  
    &  $3.93_{0.32}$  
    & $12.42_{0.73}$  
    & $21.96_{0.81}$   \\
    &&&\texttt{GPT-3.5-Turbo} & k=3 
    & $86.77_{0.09}$   
    & $18.59_{0.77}$   
    &  $5.00_{0.43}$   
    & $14.36_{0.53}$  
    & $22.44_{0.53}$ \\
    &&&\texttt{GPT-3.5-Turbo} & k=5 
    & $87.09_{0.15}$  
    & $19.86_{0.70}$   
    &  $5.56_{0.45}$  
    & $15.61_{0.73}$  
    & $22.88_{0.36}$   \\
    \cmidrule{3-10}
    &&\multirow{5}{*}{Targets} & \cellcolor{lightgray2}\citet{Syed2021GeneratingIC}& \cellcolor{lightgray2}- &
    \cellcolor{lightgray2}$89.18$ & 
    \cellcolor{lightgray2}$30.58$ & 
    \cellcolor{lightgray2}$15.73$ & 
    \cellcolor{lightgray2}$27.71$ & 
    \cellcolor{lightgray2}$26.28$
    \\&&&\texttt{GPT-3.5-Turbo} & k=0 & $85.68$&	$14.69$&	$3.61$&	$11.03$&	$22.17$  \\
    &&& \texttt{GPT-3.5-Turbo} &k=1
    & $86.67_{0.19}$
    & $18.55_{0.74}$
    &  $4.83_{0.55}$
    & $14.14_{0.67}$  
    & $22.47_{0.87}$ \\
    &&& \texttt{GPT-3.5-Turbo} &k=3 
    & $86.94_{0.27}$
    & $19.32_{1.17}$
    &  $5.24_{0.72}$
    & $15.00_{0.96}$
    & $21.88_{0.77}$ \\
    &&& \texttt{GPT-3.5-Turbo} &k=5 
    & $87.14_{0.32}$
    & $19.83_{1.22}$
    &  $5.56_{0.61}$
    & $15.58_{0.95}$
    & $21.76_{0.78}$\\
    \cmidrule{3-10}
    &&\multirow{5}{*}{Topic} & 
    \cellcolor{lightgray2}\citet{Syed2021GeneratingIC}& \cellcolor{lightgray2}- &
    \cellcolor{lightgray2}$89.38$ & 
    \cellcolor{lightgray2}$32.34$ & 
    \cellcolor{lightgray2}$17.42$ & 
    \cellcolor{lightgray2}$29.45$ & 
    \cellcolor{lightgray2}$28.22$\\
    &&& \texttt{GPT-3.5-Turbo} &k=0 & $85.75$&	$15.08$&	$3.53$&	$11.35$&	$22.31$  \\
    &&& \texttt{GPT-3.5-Turbo} &k=1
    & $86.78_{0.52}$
    & $18.47_{1.59}$
    &  $5.13_{1.22}$
    & $14.47_{1.92}$
    & $21.80_{0.93}$ \\
    &&& \texttt{GPT-3.5-Turbo} &k=3 
    & $87.14_{0.18}$
    & $19.87_{1.17}$ 
    &  $5.69_{0.86}$ 
    & $15.72_{0.99}$
    & $22.28_{1.43}$ \\
    &&& \texttt{GPT-3.5-Turbo} &k=5 
    & $87.42_{0.22}$
    & $20.63_{1.35}$  
    &  $6.14_{1.00}$
    & $16.48_{1.11}$
    & $21.90_{1.47}$ \\
    \cmidrule{2-10}
    &\multirow{10}{*}{DebateSum} &
    \multirow{5}{*}{Abstractive} & \cellcolor{lightgray2}T5-base & 
    \cellcolor{lightgray2}- &
    \cellcolor{lightgray2}$82.88$ & 
    \cellcolor{lightgray2}$11.39$ & 
    \cellcolor{lightgray2}$1.65$ & 
    \cellcolor{lightgray2}$10.41$ & 
    \cellcolor{lightgray2}$6.00$\\
    && & \texttt{GPT-3.5-Turbo} &k=0
    & $84.25$&	$10.35$&	$2.06$&	$8.28$&	$16.25$  \\
    &&& \texttt{GPT-3.5-Turbo} &k=1
    & $84.52_{0.24}$
    & $11.45_{0.71}$ 
    &  $2.24_{0.24}$
    &  $9.08_{0.60}$  
    & $16.61_{0.50}$ \\
    &&& \texttt{GPT-3.5-Turbo} &k=3 
    & $84.74_{0.20}$ 
    & $12.12_{0.55}$  
    &  $2.34_{0.19}$   
    &  $9.58_{0.51}$  
    & $16.68_{0.20}$ \\
    &&& \texttt{GPT-3.5-Turbo} &k=5 
    & $84.72_{0.17}$
    & $12.07_{0.42}$  
    &  $2.32_{0.06}$  
    &  $9.56_{0.35}$
    & $16.75_{0.05}$ \\
    \cmidrule{3-10}
    &&\multirow{5}{*}{Extractive} & \cellcolor{lightgray2}\citet{Roush2020DebateSumAL}& \cellcolor{lightgray2}- &
    \cellcolor{lightgray2}$85.90$ & 
    \cellcolor{lightgray2}$59.06$ & 
    \cellcolor{lightgray2}$44.37$ & 
    \cellcolor{lightgray2}$57.48$ & 
    \cellcolor{lightgray2}$56.55$\\
    &&& \texttt{GPT-3.5-Turbo} &k=0 & $88.36$&	$49.76$&	$30.88$&	$37.89$&	$40.62$ \\
    &&& \texttt{GPT-3.5-Turbo} &k=1
    & $88.84_{0.22}$ 
    & $51.91_{1.27}$  
    & $34.51_{1.81}$ 
    & $41.19_{1.98}$
    & $41.85_{1.58}$ \\
    &&& \texttt{GPT-3.5-Turbo} &k=3 
    & $89.52_{0.22}$
    & $55.33_{1.35}$
    & $41.05_{2.30}$  
    & $46.80_{1.88}$
    & $47.07_{2.42}$ \\
    &&& \texttt{GPT-3.5-Turbo} &k=5 
    & $89.43_{0.16}$ 
    & $54.99_{0.87}$ 
    & $40.09_{1.88}$ 
    & $45.95_{1.44}$ 
    & $46.02_{1.52}$ \\
    \bottomrule
    \end{tabular}
    }
    \caption{Performance of \texttt{GPT-3.5-Turbo} on argument generation tasks. The results are averaged over 3 random seeds for all few-shot experiments.}
    \label{tab:few-shot-AG} 
\end{table*}

Figure \ref{fig: task_difficulty} shows the effects of increasing shots on different models and task difficulties. 
In general, while there remain certain gaps between the few-shot performance of LLMs and finetuned PLMs using the full training set, it is worth noting a significant trend: by simply prompting LLMs with less than 10 demonstrations, they are able to close the gaps and match the performance of finetuned PLMs trained with 500 samples.

Comparing among models, we notice that the choice of model is crucial, as different models exhibit varying levels of proficiency across different tasks. While Flan models excel in simple tasks, the performances of \texttt{Flan-T5-XL} and \texttt{Flan-UL2} lag behind that of \texttt{GPT-3.5-Turbo} in hard tasks, even with an increased number of shots. Overall, \texttt{Flan-T5-XXL} appears to be most robust, consistently demonstrating strong performance across diverse tasks.

Secondly, larger models are not necessarily superior to smaller models.
Upon comparing the two LLaMA models, \texttt{Llama-2-13B} generally outperforms \texttt{Llama-2-7B}. However, one interesting exception surfaces when the input is minimal. Surprisingly, \texttt{Llama-2-7B} proves to be more effective than \texttt{Llama-2-13B} in simple tasks. For Flan models, larger models consistently outperform their smaller counterparts in simple tasks. This trend, however, does not hold in the case of more challenging tasks. Notably, \texttt{Flan-T5-XL} (3B) model performs comparably to \texttt{Flan-UL2} (20B) in hard tasks, despite its significantly smaller size. Furthermore, the 11B \texttt{Flan-T5-XXL} model showcases strong performance, even though it is smaller than both \texttt{Flan-UL2} and \texttt{Llama-2-13B}. This suggests that, for certain complex tasks, the performance of the model may not be solely determined by its size.

Furthermore, increasing demonstrations have varying effects on different models. \texttt{GPT-3.5-Turbo} generally benefits from more shots. For Flan models, the gain in performance is not obvious. Llama models, on the other hand, exhibit mixed performance in response to more demonstrations. The larger model demonstrates notable performance improvement, particularly in hard tasks, when provided with more shots.
However, the smaller model does not exhibit performance gain from additional demonstrations. In fact, when it comes to simple tasks, providing more shots has a negative impact. It appears that 
longer contexts might introduce noise or unnecessary information that could potentially hinder the performance of smaller models.

\subsection{Results on Argument Generation}
\label{sec:argu_gen_analysis}
Table \ref{tab:few-shot-AG} presents the performance of \texttt{GPT-3.5-Turbo} on argument generation tasks. Compared to existing SOTA, \texttt{GPT-3.5-Turbo} already outperforms previous methods in several tasks including CounterArguGen, ConcluGen in the base setting, as well as abstractive summarization in DebateSum.

Although previous methods excel in other ConcluGen settings, we attribute their high performance to additional annotations encoding specific aspects, targets, or topics. Such manual annotations are task-specific and extremely costly. \texttt{GPT-3.5-Turbo}, on the other hand, achieves comparable results across different settings regardless of the presence of encoded information.
Furthermore, the contrasting results from different evaluation metrics reveal an interesting pattern: the R\textsc{ouge} scores are generally low but the BERTScores are high. The low R\textsc{ouge} scores indicate that there are only a few overlaps between the generated text and the reference text. 
The high BERTScore indicates that the semantic meaning of the generated text is highly similar to the reference text. This suggests that although the generated text may not match the reference text in terms of exact wording or specific phrases, it successfully captures the underlying semantic meaning.
To further support this, we provide several illustrative examples in Appendix \ref{sec:appen_examples}. Both automatic evaluation and quality analysis show that \texttt{GPT-3.5-Turbo} grasps the essence of the content and conveys it effectively, even if the choice of words or phrasing differs from the reference.

For extractive summarization, the previous method \cite{Roush2020DebateSumAL} relies on word-level classification, wherein each word is predicted as either ``underlined'' or ``not-underlined'', which is inefficient and compromises the coherence of the generated sentences. In contrast, \texttt{GPT-3.5-Turbo} avoids the high training cost and generates coherent sentences. Additionally, our quality analysis shows that \texttt{GPT-3.5-Turbo} can identify important information accurately. Case studies can be found in Appendix \ref{sec:appen_ext_summ}.

In addition, we notice that \texttt{GPT-3.5-Turbo} exhibits incremental performance improvements as the number of shots increases. The performance gains are relatively modest compared to those observed in argument mining tasks. This implies that \texttt{GPT-3.5-Turbo} is inherently proficient in argument generation without necessitating more demonstrations.

We also evaluate other models including \texttt{Llama-2-7B}, \texttt{Llama-2-13B}, and \texttt{Flan-UL2} which could accommodate long context. Performance of other models are available in Appendix \ref{sec:AG}. All models exhibit similar trends with the above except \texttt{Flan-UL2} - its advantage in extractive summarization is less apparent compared to the other models.

\begin{table}[!t]
    \centering
    \resizebox{1\linewidth}{!}{
    \setlength{\tabcolsep}{0.9mm}{
    \begin{tabular}{lccccc}
    \toprule
    \textbf{Method} & \textbf{BERTScore} & \textbf{R\textsc{ouge}1} & \textbf{R\textsc{ouge}2} & \textbf{R\textsc{ouge}L} & \textbf{METEOR} \\
    \midrule
    Pipeline (Claims) & $80.33_{0.08}$  &  $31.00_{0.54}$ &     $3.70_{0.24}$ & $13.26_{0.14}$  &  $21.92_{1.21}$ \\
    \midrule
    Pipeline (Summary) & $82.23_{0.06}$&    $23.73_{6.19}$  &   $4.28_{0.92}$ &   $11.60_{2.06}$  &  $10.89_{3.41}$  \\
    \midrule
    End-to-end & $82.51_{0.05}$ &  $30.10_{1.08}$   &  $5.70_{0.18}$ &   $13.65_{0.22}$  &  $14.48_{0.77}$ \\
    \bottomrule
    \end{tabular}
    }}
    \caption{Automatic evaluation results of counter speech generation. The average scores are calculated based on three distinct sets of prompts to account for the potential sensitivity of zero-shot performance to prompt designs.}
    \label{tab:benchmark_res} 
\end{table}

\subsection{Results on Counter Speech Generation}

\paragraph{Automatic Evaluation}
Table \ref{tab:benchmark_res} shows the results from automatic evaluation. 
The end-to-end approach surpasses the summarization pipeline approach across all metrics. This highlights the model's strong capability of internalizing and synthesizing information from the supporting speech without the need for intermediate steps. 

Comparing the end-to-end approach to the claim detection pipeline approach, the former lags behind in R\textsc{ouge}-1 and METEOR, but surpasses in BERTScore, R\textsc{ouge}-2, and R\textsc{ouge}-L. To determine which approach is superior, we conduct human evaluation for a more complete understanding of the performance of these two approaches.

\paragraph{Human Evaluation}
We hired 2 human judges who are professional English speakers to manually evaluate the quality of counter speeches generated by the claim pipeline approach and the end-to-end approach on 50 random samples.
For each test instance, we provide the judges with supporting speeches along with randomly ordered counter speeches from the two methods, and ask the judges to individually evaluate the generation quality based on the following criteria:
\squishlist
\item{\textit{Fluency (Flu.)}}: Is the generation fluent, grammatical, and without unnecessary repetitions?
\item{\textit{Persuasiveness (Per.)}}: Is the text able to convince you to adopt a certain belief or attitude? 
\item{\textit{\% of arguments addressed (\% Arg.)}}: Does the counter speech address all claims/arguments in the supporting speech?
\squishend
Fluency and persuasiveness, graded on a scale of 1 to 5, are to assess the model's argument generation capability. To evaluate the model's argument mining ability, we use \% of arguments addressed, calculated by the number of addressed arguments in the counter speech over the total number of arguments in the supporting speech. This metric reflects how effectively the model can identify arguments, either explicitly in the two-step approach or implicitly in the one-step approach.

\begin{table}[!t]
    \centering
    \resizebox{0.8\linewidth}{!}{
    \setlength{\tabcolsep}{5mm}{
    \begin{tabular}{lccc}
    \toprule
    \textbf{Method} & \textbf{\textit{Flu.}} & \textbf{\textit{Per.}} & \textbf{\textit{\% Arg.}} \\
    \midrule
    Pipeline (Claims) & 3.56 & 2.8 & 78\% \\
    End-to-end & 4.32 & 3.8 & 95\% \\
    \bottomrule
    \end{tabular}
    }}
    \caption{Human evaluation scores 
on 50 test samples.}
    \label{tab:human} 
\end{table}

In Table \ref{tab:human}, it is evident that the end-to-end approach outperforms the pipeline approach on all 3 metrics.
In specific, the pipeline approach is not able to address as many arguments as the end-to-end approach, possibly due to the potential loss of information during the intermediate step. In the pipeline method, information from the supporting speech undergoes processing, such as summarization or claim detection, before the final counter speech is generated. This might result in information loss or distortion, which could negatively impact the overall coherence and effectiveness of the generated response, which in turn affects the fluency and persuasiveness scores.
In contrast, the one-step approach bypasses the intermediate stage, allowing the model to directly engage with the supporting speech and generate a counter speech in a more holistic manner. We show qualitatively in Appendix \ref{sec:case}.

\section{Conclusion and Broader Impacts}
In this paper, we have made several significant contributions to the field of computational argumentation research.

Firstly, our efforts in organizing the diverse landscape of argumentation-related tasks and standardizing the format of related datasets are crucial for future research in designing domain-specific large-scale models for argumentation.

Secondly, we for the first time systematically evaluate the performance of multiple computational argumentation tasks using LLMs in zero-shot and few-shot settings. Traditional approaches for computational argumentation rely heavily on supervised fine-tuning that requires a large amount of labeled data, hindering the progress of research in this field. Our exploration of low-resource settings addresses a gap in previous computational argumentation research, demonstrating the potential of LLMs in scenarios with limited training data.

Furthermore, we introduce a new counter speech generation benchmark that evaluates models' capability in both argument mining and argument generation. Our extensive experimental results and analysis demonstrate the potential of LLMs in computational argumentation, while also highlighting existing limitations in evaluating computational argumentation tasks.

Overall, our paper provides important insights and valuable resources for researchers interested in the field of computational argumentation, which will potentially inspire further advancement in this exciting area.

\section*{Limitation}

In the field of computational argumentation, there are more tasks involved. In this work, we only cover argument mining and argument generation, as these two categories are the most fundamental in this field. By understanding and establishing the performance of these core tasks, we could progress to tackle other tasks in our future study. 
In addition, it is challenging and laborious to conduct human evaluation on the full argument generation datasets. To address this, we could use GPT-4 as an evaluator \cite{liu-etal-2023-g}.
\section*{Acknowledgments}
This work was supported by DAMO Academy through DAMO Academy Research Intern Program. We would like to thank Wenxuan Zhang for providing feedback on the paper.

\bibliography{anthology,custom}

\clearpage
\appendix

\section{More Background}
\label{sec:appen_background}

\subsection{Argument Mining}

Argument mining is a rapidly emerging field of NLP that aims to automatically identify and extract arguments and their components from textual data. 
With the increasing volume of digital text available online, the need for automated methods to analyze and understand arguments has become more pressing. 
By identifying the arguments in natural language text, researchers can better understand the underlying beliefs, values, and motivations that drive human behavior. 
As such, argument mining is a core task of research within NLP that is poised to make significant contributions to a wide range of fields.

\subsection{Argument Generation}

With the understanding of the argumentative structures within the text through argument mining, the next step is to explore how to generate arguments.
Argument generation and argument summarization are two related tasks within computational argumentation that have the potential to transform the way we create and consume arguments.
Argument generation involves the automatic creation of persuasive text, such as generating a sentence attacking another standpoint, that can be used to influence a group of readers. 
Argument summarization, on the other hand, involves the automatic summarization of arguments, enabling users to quickly and easily understand complex arguments without having to read through lengthy documents.
For example, in the law domain, large amounts of legal documents need to be analyzed and understood in a time-sensitive manner.
As such, argument generation and summarization are two key areas of research within NLP that have the potential to significantly streamline the process of argumentation in various domains.

\section{Data Sample on Counter Speech Generation}
\label{sec:sample}

Table \ref{tab:sample} shows a data sample from our benchmark dataset for the proposed counter speech generation task.
The topic is ``Nationalism does more harm than good''.
The supporting speech is the input, and the counter speech written by humans is considered the output.

\begin{table*}[ht]
	\centering
	\resizebox{0.95\linewidth}{!}{
	    \setlength{\tabcolsep}{0mm}{
            \begin{tabular}{p{2.5cm}@{~} @{~}p{20cm}}
            \toprule
            Topic & Nationalism does more harm than good \\
            \midrule
                
            Supporting Speech

            &
            Nationalism does more harm than good. What's important to recognize about nationalism right at the outset is that it doesn't arise from anything natural about the peoples that express nationalist attitudes. There's nothing about german nationalists or french nationalists or chinese nationalists that makes those types of groups uniquely combined to each other and in fact most of these groups grew out of a very distinct cultural subsections prior to the eighteenth century. For example in germany there was no german state prior to the eighteenth century. It was a conglomeration of many different german and frankish kingdoms that came together to form a modern state, and the modern state is about when these attitudes eventually arose within our society. So it's important to recognize that there's nothing fundamentally human about nationalism, there's nothing that combines these populations in any unique way. Between the fact that they neighbor each other and in some instances share cultural bonds though when you allow for nationalism and when nationalism arises in the way that it has in the last two centuries, it allows for new different cultural bonds to be formed which are frankly exclusive in many ways and most importantly arbitrary in their creation. They're simply made in order to enforce this idea of national identity and national community that doesn't exist and is often a tool of those empowered by nationalism to use that nationalism as a guise for fascism. But firstly, before I get on to that I think it's important to talk about why nationalism is simply a bad political force within the world. Nationalism by its definition is exclusionary. In order to celebrate a nation you must create distinctions between that nation and those around it and while some would argue for a cosmopolitan nationalism that allows for people to celebrate their nation simply because it's something that is diverse and beautiful, such as the united states and the idea of the melting pot, firstly, this isn't how nationalism actually arises in the world. Nationalism is more often in more often the case, nationalism is the force that says: my national identity group, my my ethnicity, my regional nation, any sort of group is is better than other groups that border me, or that there's something that makes them distinct that makes them superior. This false superiority creates a a sense of xenophobia throughout the world, which is one reason why there's, in the, in europe right now there's such a hesitancy to to accept refugees from syria, and from other war torn areas in the middle east and northern and northern africa. This is because there's this idea that there's some sort of benefits that we read from our nation that are exclusive the benefits for our nation. That because we are where we are we have earned the goods and resources that we get from these regions. But we only get these benefits because of the arbitrary nature of where we were born and what our region happens to have and what it can give us. There's no one more deserving of getting these sort of political goods whether it be a stable government or representative democracy than people that are fleeing to these areas as refugees. It's just the luck of where they were born. Given that this is the case we think that nationalism becomes an exclusionary political philosophy that only harms the most disenfranchised people like refugees, who are not able to access the goods that they desperately need. We also think that it creates divisions within a society itself. It means that people that have become part of this communities, say minority groups in in largely white european countries, feel excluded from their own society. Whether it's through ideas of nationalism that simply don't create an image of the nation that includes them, or it's more overt and direct threats. That come from largely far right groups that use nationalism as a guise for fascism. And this the other problem with nationalism. It's that when you create xenophobic senses within a state that creates this sort of false superiority that my nation is better than your nation, it allows for strong man leaders to stand up and say: I'm going to protect the nation. I'm going to ensure the nation rises to its former glory, and these sort of robust senses of pride in the nation allow for these people to get away with crimes and other sorts of corruption that allow them to enrich themselves while at the same time creating strong men groups that create serious threats to democracy not just in developed but also in not just in developing nations but also developed nations such as greece where the xri'si party is rising, and france with marine la pen, in england with braxit and with united states and donald trump. All of these people use nationalism as a way to try and fuel their political anger that their people feel and it only creates more divisions within our society which is frankly contradictory to the global ideas that have been set forth for the past for the past sixty or seventy years of post world war two, peace and prosperity that's occurred. For these reasons we think that nationalism has certainly done more harm than good.
            \\
            \toprule

            Human Counter Speech
            &
            
            In order to consider, whether nationalism does more harm than good, you must consider the counterfactual: what would have been here had we not had nationalism? We think that, this debate is inherently comparative, in that we think, human beings have an inherent need and desire to group around things that unite them and join them together. This is why in the entire history of mankind, man has always grouped together over certain ideologies, aspects, or whatever it is. Historically, it has taken the form of religion, of monarchism, and of nationalism. Of these alternatives, we think nationalism is by far the best, and we think these alternatives are, in fact, the other options for how life may be. Let's get into rebuttal first. So first, tim says nationalism is exclusionary to other groups, he is correct about that, and then he takes it from that, and says that's why there's xenophobia, that is what he is incorrect about. Xenophobia existed far before nationalism. Religions fought amongst themselves for millennia, so did monarchies who went to war over crown crown and queen, for example. We don't think nationalism caused that. In fact, we think since the rise of nationalism, national wars have gone drastically down. Secondly, he says: minority groups within society feel exclusion, excluded. Again, let's look comparatively. We think a jew, in a christian society, is inherently excluded from that society. We think, an israeli in a american society, can take upon himself aspects of american nationalism, without giving up his religious identity, and thereby allow him to participate in society, more than other groupings would. Lastly, he says: it allows for corrupt leaders. We accept this, it's true. We think it's less so than the alternatives, that are based on a deity. Let's take a look into that. Why is nationalism better? Two reasons: one, based on leaders, second, based on geographic inclusion. First, let's talk about leaders. We think what makes nationalism unique, is that it puts the people in the middle. The comparative of nationalism is various forms of identity, that all include one central leader, be it god, be it chief rabbi, be it a king or a queen. We think that is particularly dangerous, because it allows for that corrupt power, in a significantly more powerful way, than any form of identity based on the nation as a whole. At the point, at which even the leader can be seen to be harming the nation, we think, that nationalism allows groups to protect themselves from corrupt leaders. It is true, that in instances, it also allows them to fall to corrupt leaders, but historically, we think you have far more corrupt leaders under alternative ways of grouping society. So, we think nationalism is better based on the leaders. Let's talk about geographic inclusion. At the point, at which you have two " otherize " some group, because in order to unite yourself with some people, it inherently necessitates creating some form of enemy, and this has been true all throughout history. We think, the best way of doing that, is uniting yourself around the group, based on where you are geographically located. We think that's better, because it's much more difficult to start wars with people who are far away from you. We think that's better, because it's much more difficult to have local tensions, if all of your enemies are far away from you. We think it's better, because all of the reasons, for which humans tend to strive to be in groups, mean that they gain more from these groups, when they are surrounded by these groups. So, nationalism is the best form of grouping together, and grouping together is inherent to human nature. For these reasons, we think nationalism has done far more good than harm.
            \\

            
            
            

            

            \bottomrule
            \end{tabular}
        }
    }
    \caption{A data sample of the benchmark dataset for the counter speech generation task.}
\label{tab:sample}
\end{table*}

\section{Prompt Templates}
\label{sec:prompts}
\subsection{Prompt Templates for Argument Mining Tasks}

Table \ref{tab:step1_prompt} shows the prompt templates for selected argument mining tasks, including claim detection and stance detection.

\begin{table*}[ht]
    \centering
    \resizebox{\columnwidth}{!}{
	\setlength{\tabcolsep}{0mm}{
            \begin{tabular}{p{10cm}}
            \toprule
            Template for claim detection \\
            \midrule
                Identify whether the given sentence is a claim towards the given topic. Choose from 'claim' or 'non claim'. \\\\
                
                Sentence: \textcolor{blue}{[sentence]}\\
                Claim: \textcolor{blue}{[claim]}\\
                Label: \\
            \toprule
            Template for stance detection \\
            \midrule
                Identify the stance of the given sentence towards each given target. Choose from 'support', 'attack', or 'neutral' for each target in the target pair. Format the output as a label pair: label1, label2. \\\\

                Sentence: \textcolor{blue}{[sentence]}\\
                Target Pair: \textcolor{blue}{[targets]}\\
                Label Pair: \\

            \bottomrule
            \end{tabular}

        }
    }
    \caption{
        Prompt templates for selected argument mining tasks.
    }
    \label{tab:step1_prompt}
\end{table*}

\subsection{Prompt Templates of Argument Generation Tasks}

\begin{table*}[ht]
    \centering
    \resizebox{\columnwidth}{!}{
	\setlength{\tabcolsep}{0mm}{
            \begin{tabular}{p{10cm}}
            \toprule
            Template for counter argument generation \\
            \midrule
                Identify a premise for a claim and come up with a counter-argument that challenges the validity of that premise. \\\\
                
                Claim: \textcolor{blue}{[claim]}\\
                Premises: \textcolor{blue}{[premises]}\\
                Counter Argument: \\
            \midrule
            Template for abstractive summarization \\
            \midrule
                Identify the main points and supporting evidence in the document that support the argument being made. \\\\

                Document: \textcolor{blue}{[document]}\\
                Abstractive Summary: \\

            \bottomrule
            \end{tabular}

        }
    }
    \caption{
        Prompt templates for selected argument generation tasks.
    }
    \label{tab:step2_prompt}
\end{table*}

Table \ref{tab:step2_prompt} shows the prompt templates for argument generation tasks, including counter argument generation and abstractive summarization.

\section{Training Details of SOTA}
\label{sec:training}
For argument mining tasks, we train sentence-pair classifiers based on pre-trained models
such as BERT \cite{Devlin2019BERTPO} following the settings reported by \citet{Cheng2022IAMAC}. Dataset statistics can be found in Table \ref{tab:stats}. For datasets where training sets are not available, we randomly sample 500 data points to reserve for the test set and make use of all the remaining for training.

For CounterArguGen, we directly evaluate based on the released predictions \cite{alshomary2021b}.

For ConcluGen \cite{Syed2021GeneratingIC}, we use the available checkpoints and conduct inference on our sampled test set. The released checkpoints were trained on the reported training sets, while our 500 samples were sampled from the test sets for a proper train-test split.

For DebateSum tasks, we randomly sample 90000 data for the train set, 10000 for the development set, and 500 for the test set, since the original train test split is not specified. Specifically, for abstractive summarization, we finetune a T5-base \cite{t5}, a popular and performant generative model, using the AdamW optimizer with a learning rate of 1e-4, a fixed batch size of 4, and 3 training epochs. For DebateSum extractive summarization, we follow the settings reported by \citet{Roush2020DebateSumAL}. 

\begin{table*}[!t]
    \centering
    \resizebox{0.7\linewidth}{!}{
    \begin{tabular}{llcccc}
    \toprule
    Task & Dataset & Train & Dev & Test & Class \\
    \midrule
    \multirow{3}{*}{Claim Detection} & \setulcolor{grass2}\ul{IBM Claims} & 1500 & 500 & 500 & 2 \\
    & \setulcolor{grass2}\ul{IBM Argument} & 99 & 9 & 500 & 2 \\
    & \setulcolor{grass2}\ul{IAM Claims} & 55044 & 500 & 500 & 2\\
    \midrule
    \multirow{2}{*}{Evidence Detection} & \setulcolor{grass2}\ul{IBM Evidence} & 3566 & 500 & 500 & 2 \\
    & \setulcolor{grass2}\ul{IAM Evidence} & 56898 & 500 & 500 & 2 \\
    \midrule
    \multirow{4}{*}{Stance Detection} & \setulcolor{grass2}\ul{IAM Stance} & 3371 & 500 & 500 & 2 \\
    & \setulcolor{grass2}\ul{IBM Stance} & 1500 & 500 & 500 & 2 \\
    & \setulcolor{myred}\ul{MTSD} & 5738 & 500 & 500 & 9 \\
    & \setulcolor{myred}\ul{FEVER} & 144949 & 500 & 500 & 3 \\
    \midrule
    \multirow{2}{*}{Evidence Classification} & \setulcolor{myred}\ul{IBM Type} & 291 & 500 & 500 & 3\\
    & \setulcolor{myred}\ul{AQE Type} & 7407 & 500 & 500 & 5 \\
    \midrule
    Generation & CounterArguGen & 0 & 0 & 100 & - \\
    \midrule
    \multirow{6}{*}{Summarization} & ConcluGen-base & 123539 & 12354 & 500 & -\\
    & ConcluGen-Aspects & 122040 & 12192 & 500 & -\\
    & ConcluGen-Targets & 110867 & 11068 & 500 & -\\
    & ConcluGen-Topic & 123538 & 12354 & 500 & -\\
    & DebateSum-Abstractive & 90000 & 10000 & 500 & -\\
    & DebateSum-Extractive & 90000 & 10000 & 500 & -\\
    \bottomrule
    \end{tabular}
    }
    \caption{Dataset statistics.}
    \label{tab:stats} 
\end{table*}

\section{Additional Results on Argument Mining}
\label{sec:appen_am_few}

\begin{table*}[!t]
    \centering
    \resizebox{0.98\linewidth}{!}{
    \begin{tabular}{lccccccccccc}
    \toprule
    \multirow{2}{*}{\textbf{Model}} &
    \multicolumn{3}{c}{\textbf{Claim Detection}} &
    \multicolumn{2}{c}{\textbf{Evidence Detection}} &
    \multicolumn{4}{c}{\textbf{Stance Detection}} &
    \multicolumn{2}{c}{\textbf{Evidence Classification}} \\
    \cmidrule(lr){2-4}
    \cmidrule(lr){5-6}
    \cmidrule(lr){7-10}
    \cmidrule(lr){11-12} 
    & \textbf{\setulcolor{grass2}\ul{IBM Claims}} & \textbf{\setulcolor{grass2}\ul{IBM Argument}} & \textbf{\setulcolor{grass2}\ul{IAM Claims}} & 
    \textbf{\setulcolor{grass2}\ul{IBM Evidence}} & 
    \textbf{\setulcolor{grass2}\ul{IAM Evidence}} & \textbf{\setulcolor{grass2}\ul{IAM Stance}} & 
    \textbf{\setulcolor{grass2}\ul{IBM Stance}} & \textbf{\setulcolor{myred}\ul{MTSD}} & 
    \textbf{\setulcolor{myred}\ul{FEVER}} & 
    \textbf{\setulcolor{myred}\ul{IBM Type}} & 
    \textbf{\setulcolor{myred}\ul{AQE Type}} \\ 
    \midrule
    \multicolumn{11}{c}{$Acc.$} \\
    \midrule
    \rowcolor{lightgray2} 
    \texttt{Random} & 50.20 & 48.00 & 52.00 & 49.20 & 51.00 & 45.80 & 47.00 & 10.00 & 29.40 & 33.20 & 19.20 \\ 
    \texttt{Flan-T5-XL} & 74.00 & 59.00 & 91.60 & 69.60 &
    77.40 & 53.80 & 27.80 & 15.40 & 34.20 & 73.80 & 22.20 \\ 
    \texttt{Flan-T5-XXL} & 72.00 & 59.20 & 88.20 & 71.60 & 82.00 & 61.60 & 38.00 & 32.00 & 34.80 & 75.20 & 56.20 \\ 
    \texttt{Llama-2-7B} & 32.40 & 41.20 & 48.00 & 45.20 &
    33.60 & 11.20 & 7.00 & 4.20 & 29.40 & 6.40 & 2.40 \\ 
    \midrule
    \multicolumn{11}{c}{$F_1$} \\
    \midrule
    \rowcolor{lightgray2} 
    \texttt{Random} & 55.53 & 52.01 & 64.28 & 49.62 & 58.05 & 51.76 & 50.05 & 12.59 & \textbf{33.93} & 33.51 & 24.42 \\
    \texttt{Flan-T5-XL} & 69.07 & 58.27 & 90.87 & 68.69 & 80.09 & 43.55 & 13.46 & 13.68 & 27.33 & 73.66 & 15.74 \\
    \texttt{Flan-UL2 (20B)} & 63.49 & 54.69 & 89.22 & 69.17 & 82.78 & 60.34 & 38.75 & 31.02 & 28.47 & 74.83 & 57.80 \\
    \texttt{Llama-2-7B} & 24.87 & 31.34 & 60.14 & 42.81 & 38.50 & 15.91 & 7.73 & 0.68 & 16.79 & 10.24 & 2.59 \\
    \bottomrule
    \end{tabular}
    }
    \caption{Zero-shot performance of \texttt{Flan-T5-XL}, \texttt{Flan-T5-XXL} and \texttt{Llama-2-7B} on argument mining tasks.}
    \label{tab:zero-shot-AM-others} 
\end{table*}

Table \ref{tab:zero-shot-AM-others} shows the zero-shot performance of \texttt{Flan-T5-XL}, \texttt{Flan-T5-XXL} and \texttt{Llama-2-7B} on argument mining tasks.

\section{Quality Analysis on Argument Generation Tasks}
\label{sec:appen_examples}

\begin{table*}[ht]
    \centering
    \resizebox{0.7\linewidth}{!}{
	\setlength{\tabcolsep}{0mm}{
            \begin{tabular}{p{2cm}@{~} @{~}p{12cm}}
            \toprule
            Reference 1 & Professional teams shouldn't be required to announce or release the name of their \colorbox{pink}{inactive players}.\\
            Prediction 1 & Teams should not be required to release a list of players that \colorbox{pink}{cannot play due to injury or other reasons}, as it takes away a strategic advantage for the team.\\
            \midrule
            Reference 2 & Free-to-play games are \colorbox{pink}{the worst thing} happening in the gaming industry today.\\
            Prediction 2 & The rise in popularity of free to play games and their associated practices such as micro transactions and pay to win will \colorbox{pink}{have a negative impact} on the gaming industry as a whole.\\
            \midrule
            Reference 3 & I believe that bicyclists \colorbox{pink}{need to} \colorbox{pink}{follow the same set of rules} that cars or motorcycles do while on the road, up to and including minimum speed, lane splitting, signaling, and traffic signs. Failing that, they need to stay off of the road.\\
            Prediction 3 & Cyclists \colorbox{pink}{should} \colorbox{pink}{be held to the same standards} as motorists when it comes to obeying traffic laws and regulations.\\
            \bottomrule
            \end{tabular}

        }
    }
    \caption{
         Examples of the references and predictions from the ConcluGen dataset. Phrases with similar meanings but different expressions are highlighted in \colorbox{pink}{pink}.
    }
    \label{tab:conclugen_examples}
\end{table*}

To further support our claims in Section \ref{sec:argu_gen_analysis}, we show three examples of references and predictions from the ConcluGen dataset in Table \ref{tab:conclugen_examples}.
For instance, in the third pair, while the generated text uses ``should'' instead of ``need to'', and ``be held to the same standards'' instead of ``follow the same set of rules'', it effectively conveys the same meaning as the reference.
These observations imply that the generated text might have used different wordings but the overall semantic meaning is similar to that of the reference text, which further supports our claims.

\section{Case Study on Extractive Summarization}
\label{sec:appen_ext_summ}

To further support our claims in Section \ref{sec:argu_gen_analysis}, we show 2 examples in Table \ref{tab:ext_case1} and \ref{tab:ext_case2}. It can be observed that Longformer tends to generate incoherent sentences, while \texttt{GPT-3.5-Turbo} can generate coherent sentences and can extract important information from the input.

\begin{table*}[ht]
	\centering
	\resizebox{\linewidth}{!}{
	    \setlength{\tabcolsep}{0mm}{
            \begin{tabular}{p{2.5cm}@{~} @{~}p{20cm}}
            \toprule
            Reference & I'm not sure a wave will necessarily mean the minority party will wrestle away control of the House. \textcolor{blue}{I'd argue a wave doesn't just need to be measured by seats won. It can be measured by votes won.} It's on this score that Democrats are in a very strong position historically speaking. The problem for Democrats isn't lack of popular support. It's how that support gets translated into seats It's not unusual historically speaking for the minority party to need more than a majority of votes (cast for the two major parties) to win a majority of seats. That's because incumbents tend to outperform the national environment, and the majority party usually has more incumbents running. What is unusual about 2018 is how great the disparity is between votes and seats is likely to be because of geographic sorting and gerrymandering. Usually, the minority party would need to win the House popular vote by a few points to win control, not seven. \textcolor{blue}{It just strikes me as unreasonable to expect that Democrats to win a net gain of much more than 23 seats given the vote/seat disparity and that no minority party in a midterm in the modern era has done better than an 8.5-point popular vote win.} If Democrats are able to do that, it would be extraordinary.
 \\
            \midrule

            Longformer
            & 
            we to I that an electoral will the of and the elections in those I'm not a wave will mean the minority party will wrestle away control of the House. is this to a the and the is a wave? \textcolor{red}{a wave doesn't just need to be by seats won. It can be by votes won.} It's on this that Democrats are in a In of over the Democrats minority party) held a 7.5-percentage point on the No minority party has won the national House vote in a election by more than the percentage won it by in the Democrats the national House vote by in The for the election is and the the minority Republicans won it by in and is, Democrats are for a minority party. The for Democrats isn't of It's how that support gets into seats. A 7.5-point win in the national House vote Democrats in the of they need to be for a of seats to win a of seats. The level of disparity between seats and votes won is from it a that about at It's not unusual for the minority party to need more than a of votes for the parties) to win a majority of seats. That's because incumbents to the national and the majority party has more incumbents is about is how the disparity is between votes and seats is likely to be because of and the minority party would need to win the House popular vote by a to win control, not It's this that makes it in to a wave on the of seats won and in terms of control of the it's seats that It's that seats should be included in of a It to that \textcolor{red}{Democrats to win a net of much more than seats the vote/seat and that no minority party in a in the has done than an popular vote win.} Democrats are able to do that, it would be

            \\

            \midrule

            \texttt{GPT-3.5-Turbo} & The article argues that a wave election in the upcoming November elections may not necessarily result in the minority party (Democrats) winning control of the House. While Democrats currently hold a 7.5-percentage point lead on the generic congressional ballot, the level of disparity between votes and seats is likely to be significant due to geographic sorting and gerrymandering. \textcolor{blue}{The article suggests that a wave should not just be measured by seats won, but also by votes won.} Democrats are doing well for a minority party, but the problem is how that support gets translated into seats. \textcolor{blue}{The article concludes that it may be difficult for Democrats to win a net gain of much more than 23 seats given the vote/seat disparity and the historical precedent of no minority party in a midterm election winning the national House vote by more than 8.5 percentage points.}

            \\
            \bottomrule
            \end{tabular}
        }
    }
    \caption{Case study 1 of extractive summarization. Incoherent sentences are in \textcolor{red}{red}. The effectively extracted important information is in \textcolor{blue}{blue}.}

\label{tab:ext_case1}
\end{table*}

\begin{table*}[ht]
	\centering
	\resizebox{\linewidth}{!}{
	    \setlength{\tabcolsep}{0mm}{
            \begin{tabular}{p{2.5cm}@{~} @{~}p{20cm}}
            \toprule
            
            Reference & 
            \textcolor{blue}{Latin American markets are becoming increasingly valuable for Chinese companies because they allow the PRC to expand and diversify its export base at a time when economic growth is slowing.} The region has also proven an effective market for Chinese efforts to sell more sophisticated, higher value added products in sectors seen as strategic, such as automobiles, appliances, computers and telecommunication equipment, and aircraft In expanding markets in Latin American the hopes of access to Chinese markets and investments among key groups of business people and government officials in those nations have played a key role in the political will to overcome the resistance \textcolor{blue}{China has applied more explicit pressures to induce Latin America to keep its markets open to Chinese goods}
 \\
            \midrule

            Longformer
            & 
            \textcolor{red}{Access to Latin American Latin American markets are becoming for Chinese companies because they allow the PRC to expand and its export base at a when economic growth is in markets} as the United and The region has also proven an effective market for Chinese efforts to sell more higher value added products in sectors as as and and In expanding access for its products through trade with as and and markets in Latin American with as and the PRC has to overcome resistance by and in those In the of access to Chinese markets and investments among of and in those have a in the political to overcome the resistance. it that the Chinese to of the in the could on the and an if an a Chinese Protection of Chinese Investments in and Trade from the \textcolor{red}{China has applied more to induce Latin America to keep its markets open to Chinese goods.} It has measures by the Argentine and that it has as protectionist: and, in the of as informal China a causing billion in soy exports and for

            \\

            \midrule

            \texttt{GPT-3.5-Turbo} 
            & 
            \textcolor{blue}{Latin American markets are becoming increasingly valuable for Chinese companies as they allow the PRC to expand and diversify its export base.} The region has also proven an effective market for Chinese efforts to sell more sophisticated, higher value-added products in strategic sectors. In expanding access for its products through free trade accords with countries such as Chile, Peru, and Costa Rica, and penetrating markets in Latin American countries with existing manufacturing sectors such as Mexico, Brazil, and Argentina, the PRC has often had to overcome resistance by organized and often politically well-connected established interests in those nations. \textcolor{blue}{China has applied more explicit pressures to induce Latin America to keep its markets open to Chinese goods,} and it has specifically protested measures by the Argentine and Mexican governments that it has seen as protectionist. In the case of Argentina, China began enforcing a longstanding phytosanitary regulation, causing almost \$2 billion in lost soy exports and other damages for Argentina.

            \\
            \bottomrule
            \end{tabular}
        }
    }
    \caption{Case study 2 of extractive summarization. Incoherent sentences are in \textcolor{red}{red}. The effectively extracted important information is in \textcolor{blue}{blue}.}

\label{tab:ext_case2}
\end{table*}

\section{Additional Results on Argument Generation}
\label{sec:AG}
Table \ref{tab:few-shot-AG-llama7b}, \ref{tab:few-shot-AM-llama13b} and \ref{tab:few-shot-AG-flan} display the evaluation results of argument generation tasks using \texttt{Llama-2-7B}, \texttt{Llama-2-13B} and \texttt{Flan-UL2} respectively.

\begin{table*}[!t]
    \centering
    \resizebox{0.95\linewidth}{!}{
    \begin{tabular}{lllllllll}
    \toprule
    Task & Dataset & Setting & k-shot & BERTScore & R\textsc{ouge}1 & R\textsc{ouge}2 & R\textsc{ouge}L & METEOR \\
    \midrule
    \multirow{2}{*}{Generation} & \multirow{2}{*}{CounterArguGen} & Premises & k=0 & $77.42$ & $6.40$ & $1.01$ & $4.98$ & $9.81$ \\
    \cmidrule{3-9}
    && Weak Premises & k=0 & $76.87$ & $5.28$ & $0.62$ & $4.76$ & $8.39$ \\
    \midrule
    \multirow{24}{*}{Summarization} &
    \multirow{16}{*}{ConcluGen} &
    \multirow{4}{*}{Base}
    & k=0 & $78.08$ & $3.23$ & $0.98$ & $2.99$ & $8.83$ \\
    &&& k=1 
    & $76.75_{0.84}$
    & $2.49_{0.45}$
    & $0.57_{0.21}$
    & $2.27_{0.35}$
    & $6.35_{1.07}$ \\
    &&& k=3
    & $76.36_{0.43}$
    & $1.93_{0.22}$
    & $0.32_{0.09}$
    & $1.82_{0.17}$
    & $5.02_{0.56}$ \\
    &&& k=5
    & $76.24_{0.27}$
    & $1.87_{0.22}$
    & $0.30_{0.09}$
    & $1.77_{0.19}$
    & $4.83_{0.46}$ \\
    \cmidrule{3-9}
    &&\multirow{4}{*}{Aspects}
    & k=0 & $78.03$ & $4.64$ & $1.41$ & $4.22$ & $10.51$ \\
    &&& k=1
    & $76.55_{1.02}$
    & $3.55_{0.14}$
    & $0.81_{0.10}$
    & $3.22_{0.08}$
    & $7.86_{0.14}$ \\
    &&& k=3
    & $76.17_{1.12}$
    & $2.43_{1.03}$
    & $0.41_{0.21}$
    & $2.22_{1.00}$
    & $5.52_{2.17}$ \\
    &&& k=5
    & $77.19_{0.34}$
    & $3.07_{0.69}$
    & $0.68_{0.27}$
    & $2.75_{0.63}$
    & $7.13_{1.24}$ \\
    \cmidrule{3-9}
    &&\multirow{4}{*}{Targets}
    & k=0 & $78.10$ & $4.35$ & $1.28$ & $3.98$ & $10.42$ \\
    &&& k=1
    & $77.49_{0.48}$
    & $3.97_{0.45}$
    & $0.98_{0.11}$
    & $3.58_{0.40}$
    & $9.09_{0.87}$ \\
    &&& k=3
    & $77.73_{1.25}$
    & $3.11_{0.45}$
    & $0.59_{0.22}$
    & $2.71_{0.37}$
    & $7.50_{1.04}$ \\
    &&& k=5
    & $77.41_{1.70}$
    & $2.91_{0.97}$
    & $0.41_{0.22}$
    & $2.54_{0.58}$
    & $6.75_{1.96}$ \\
    \cmidrule{3-9}
    &&\multirow{5}{*}{Topic}
    & k=0 & $77.29$ & $3.78$ & $1.16$ & $3.51$ & $9.13$ \\
    &&& k=1
    & $77.22_{0.58}$
    & $3.23_{0.33}$
    & $0.71_{0.13}$
    & $2.97_{0.26}$
    & $7.38_{1.11}$ \\
    &&& k=3
    & $76.75_{0.46}$
    & $2.20_{0.57}$
    & $0.34_{0.20}$
    & $2.02_{0.54}$
    & $5.41_{1.40}$ \\
    &&& k=5
    & $76.83_{0.81}$
    & $2.09_{0.98}$
    & $0.29_{0.23}$
    & $1.87_{0.82}$
    & $5.27_{2.25}$ \\
    \cmidrule{2-9}
    & \multirow{8}{*}{DebateSum} &
    \multirow{4}{*}{Abstractive}
    & k=0 & $78.55$ & $3.14$ & $0.61$ & $2.71$ & $7.32$ \\
    &&& k=1
    & $77.83_{0.60}$
    & $2.72_{0.34}$
    & $0.51_{0.08}$
    & $2.35_{0.31}$
    & $6.30_{0.83}$ \\
    &&& k=3
    & $77.93_{0.49}$
    & $2.72_{0.33}$
    & $0.50_{0.09}$
    & $2.33_{0.31}$
    & $6.33_{0.77}$ \\
    &&& k=5
    & $77.89_{0.53}$
    & $2.73_{0.32}$
    & $0.52_{0.08}$
    & $2.36_{0.30}$
    & $6.37_{0.75}$ \\
    \cmidrule{3-9}
    &&\multirow{4}{*}{Extractive}
    & k=0 & $83.71$ & $34.45$ & $24.90$ & $29.74$ & $41.47$ \\
    &&& k=1
    & $84.60_{0.72}$
    & $36.90_{2.08}$
    & $27.59_{2.27}$
    & $31.43_{1.62}$
    & $44.50_{2.59}$ \\
    &&& k=3
    & $83.89_{0.84}$
    & $34.70_{2.24}$
    & $25.13_{2.96}$
    & $29.47_{2.28}$
    & $41.38_{3.97}$ \\
    &&& k=5
    & $84.91_{0.94}$
    & $37.99_{2.68}$
    & $28.96_{3.15}$
    & $32.73_{2.41}$
    & $45.97_{3.56}$ \\
    \bottomrule
    \end{tabular}
    }
    \caption{Performance on argument generation tasks using \texttt{Llama-2-7B}.}
    \label{tab:few-shot-AG-llama7b}
\end{table*}

\begin{table*}[!t]
    \centering
    \resizebox{0.95\linewidth}{!}{
    \begin{tabular}{lllllllll}
    \toprule
    Task & Dataset & Setting & k-shot & BERTScore & R\textsc{ouge}1 & R\textsc{ouge}2 & R\textsc{ouge}L & METEOR \\
    \midrule
    \multirow{2}{*}{Generation} & \multirow{2}{*}{CounterArguGen} & Premises & k=0 & $78.01$ & $7.53$ & $0.70$ & $5.59$ & $11.29$ \\
    \cmidrule{3-9}
    && Weak Premises & k=0 & $78.06$ & $7.87$ & $0.91$ & $5.95$ & $11.85$ \\
    \midrule
    \multirow{24}{*}{Summarization} &
    \multirow{16}{*}{ConcluGen} &
    \multirow{4}{*}{Base}
    & k=0 & $78.99$ & $3.95$ & $1.10$ & $3.49$ & $9.73$ \\
    &&& k=1 
    & $76.05_{0.43}$
    & $2.79_{0.18}$
    & $0.55_{0.10}$
    & $2.42_{0.15}$
    & $6.55_{0.65}$ \\
    &&& k=3
    & $76.63_{0.02}$
    & $2.55_{0.09}$
    & $0.45_{0.00}$
    & $2.18_{0.04}$
    & $6.18_{0.21}$ \\
    &&& k=5
    & $77.39_{0.36}$
    & $2.87_{0.08}$
    & $0.57_{0.03}$
    & $2.36_{0.05}$
    & $7.17_{0.31}$ \\
    \cmidrule{3-9}
    &&\multirow{4}{*}{Aspects}
    & k=0 & $78.47$ & $4.04$ & $1.18$ & $3.54$ & $9.68$ \\
    &&& k=1
    & $77.38_{0.40}$
    & $3.90_{0.23}$
    & $0.83_{0.08}$
    & $3.38_{0.22}$
    & $8.57_{0.22}$ \\
    &&& k=3
    & $77.33_{0.25}$
    & $3.59_{0.37}$
    & $0.69_{0.13}$
    & $3.09_{0.34}$
    & $7.95_{0.73}$ \\
    &&& k=5
    & $77.36_{0.64}$
    & $3.84_{0.17}$
    & $0.84_{0.00}$
    & $3.37_{0.17}$
    & $8.61_{0.13}$ \\
    \cmidrule{3-9}
    &&\multirow{4}{*}{Targets}
    & k=0 & $78.58$ & $4.25$ & $1.27$ & $3.79$ & $10.37$ \\
    &&& k=1
    & $78.14_{0.33}$
    & $4.38_{0.33}$
    & $1.19_{0.11}$
    & $3.95_{0.17}$
    & $10.17_{0.66}$ \\
    &&& k=3
    & $78.42_{0.49}$
    & $3.96_{0.69}$
    & $0.91_{0.44}$
    & $3.42_{0.62}$
    & $9.16_{1.40}$ \\
    &&& k=5
    & $77.69_{0.34}$
    & $3.10_{0.68}$
    & $0.64_{0.24}$
    & $2.71_{0.67}$
    & $7.54_{1.10}$ \\
    \cmidrule{3-9}
    &&\multirow{5}{*}{Topic}
    & k=0 & $78.71$ & $3.91$ & $1.11$ & $3.48$ & $9.66$ \\
    &&& k=1
    & $77.80_{0.52}$
    & $3.83_{0.31}$
    & $1.01_{0.11}$
    & $3.50_{0.26}$
    & $9.08_{0.83}$ \\
    &&& k=3
    & $77.92_{0.77}$
    & $3.18_{0.40}$
    & $0.59_{0.08}$
    & $2.72_{0.28}$
    & $7.62_{0.73}$ \\
    &&& k=5
    & $77.68_{0.81}$
    & $2.98_{0.66}$
    & $0.58_{0.20}$
    & $2.55_{0.57}$
    & $7.40_{1.34}$ \\
    \cmidrule{2-9}
    & \multirow{8}{*}{DebateSum} &
    \multirow{4}{*}{Abstractive}
    & k=0 & $78.97$ & $3.35$ & $0.62$ & $2.90$ & $7.58$ \\
    &&& k=1
    & $78.39_{0.67}$
    & $2.98_{0.49}$
    & $0.51_{0.16}$
    & $2.50_{0.40}$
    & $6.63_{1.15}$ \\
    &&& k=3
    & $78.57_{0.35}$
    & $2.98_{0.37}$
    & $0.53_{0.15}$
    & $2.51_{0.36}$
    & $6.75_{0.91}$ \\
    &&& k=5
    & $78.38_{0.95}$
    & $3.09_{0.22}$
    & $0.55_{0.10}$
    & $2.59_{0.22}$
    & $6.71_{0.96}$ \\
    \cmidrule{3-9}
    &&\multirow{4}{*}{Extractive}
    & k=0 & $83.33$ & $32.59$ & $21.45$ & $26.62$ & $37.57$ \\
    &&& k=1
    & $84.29_{1.04}$
    & $35.45_{3.25}$
    & $25.08_{3.95}$
    & $29.09_{3.23}$
    & $41.43_{4.32}$ \\
    &&& k=3
    & $83.74_{0.77}$
    & $33.63_{1.51}$
    & $22.74_{2.13}$
    & $27.07_{1.37}$
    & $38.60_{2.70}$ \\
    &&& k=5
    & $84.65_{1.15}$
    & $37.03_{3.57}$
    & $26.81_{4.58}$
    & $30.72_{3.32}$
    & $43.16_{4.55}$ \\
    \bottomrule
    \end{tabular}
    }
    \caption{Performance on argument generation tasks using \texttt{Llama-2-13B}.}
    \label{tab:few-shot-AM-llama13b}
\end{table*}

\begin{table*}[!t]
    \centering
    \resizebox{0.95\linewidth}{!}{
    \begin{tabular}{lllllllll}
    \toprule
    Task & Dataset & Setting & k-shot & BERTScore & R\textsc{ouge}1 & R\textsc{ouge}2 & R\textsc{ouge}L & METEOR \\
    \midrule
    \multirow{2}{*}{Generation} & \multirow{2}{*}{CounterArguGen} & Premises & k=0 & $84.35$ & $10.38$ & $1.03$ & $8.01$ &	$5.64$  \\
    \cmidrule{3-9}
    && Weak Premises & k=0 & $84.39$ & $11.76$ & $1.60$ & $8.71$ &	$6.76$  \\
    \midrule
    \multirow{24}{*}{Summarization} &
    \multirow{16}{*}{ConcluGen} &
    \multirow{4}{*}{Base}
    & k=0 & $87.37$ & $21.54$ & $8.19$ & $19.03$ & $14.92$ \\
    &&& k=1 
    & $87.84_{0.02}$
    & $23.45_{0.10}$
    & $8.86_{0.22}$
    & $20.61_{0.08}$
    & $16.82_{0.35}$ \\
    &&& k=3
    & $87.92_{0.05}$
    & $23.94_{0.27}$
    & $9.06_{0.50}$
    & $21.00_{0.36}$
    & $17.22_{0.64}$ \\
    &&& k=5
    & $87.99_{0.08}$
    & $24.23_{0.14}$
    & $9.26_{0.32}$
    & $21.17_{0.15}$
    & $17.60_{0.59}$ \\
    \cmidrule{3-9}
    &&\multirow{4}{*}{Aspects}
    & k=0 & $87.41$ & $22.39$ & $8.86$ & $19.81$ & $16.82$ \\
    &&& k=1
    & $87.54_{0.12}$
    & $22.95_{0.52}$
    & $8.62_{0.06}$
    & $19.79_{0.29}$
    & $19.14_{0.85}$ \\
    &&& k=3
    & $87.73_{0.09}$
    & $23.83_{0.19}$
    & $8.97_{0.26}$
    & $20.44_{0.25}$
    & $20.72_{0.41}$ \\
    &&& k=5
    & $87.63_{0.05}$
    & $23.34_{0.27}$
    & $8.71_{0.16}$
    & $20.07_{0.29}$
    & $19.89_{0.34}$ \\
    \cmidrule{3-9}
    &&\multirow{4}{*}{Targets}
    & k=0 & $87.40$ & $22.58$ & $8.65$ & $19.89$ & $17.06$ \\
    &&& k=1
    & $87.62_{0.09}$
    & $23.46_{0.39}$
    & $8.90_{0.25}$
    & $20.33_{0.31}$
    & $19.69_{1.26}$ \\
    &&& k=3
    & $87.61_{0.12}$
    & $23.38_{0.21}$
    & $8.95_{0.13}$
    & $20.22_{0.22}$
    & $19.85_{0.68}$ \\
    &&& k=5
    & $87.63_{0.13}$
    & $23.30_{0.48}$
    & $8.81_{0.19}$
    & $20.13_{0.40}$
    & $19.99_{0.84}$ \\
    \cmidrule{3-9}
    &&\multirow{4}{*}{Topic}
    & k=0 & $87.59$ & $22.50$ & $8.32$ & $19.84$ & $16.53$ \\
    &&& k=1
    & $87.82_{0.17}$
    & $23.83_{0.64}$
    & $8.86_{0.40}$
    & $20.81_{0.31}$
    & $18.76_{1.62}$ \\
    &&& k=3
    & $87.90_{0.04}$
    & $24.21_{0.21}$
    & $8.88_{0.39}$
    & $20.83_{0.33}$
    & $19.70_{0.58}$ \\
    &&& k=5
    & $87.92_{0.09}$
    & $24.02_{0.24}$
    & $8.89_{0.54}$
    & $20.85_{0.44}$
    & $19.52_{1.03}$ \\
    \cmidrule{2-9}
    & \multirow{8}{*}{DebateSum} &
    \multirow{4}{*}{Abstractive}
    & k=0 & $85.39$ & $14.97$ & $2.70$ & $12.55$ & $12.73$ \\
    &&& k=1
    & $83.81_{2.75}$
    & $11.42_{5.76}$
    & $2.06_{1.09}$
    & $9.55_{4.78}$
    & $9.76_{4.74}$ \\
    &&& k=3
    & $85.45_{0.11}$
    & $14.92_{0.14}$
    & $2.80_{0.08}$
    & $12.53_{0.11}$
    & $12.57_{0.07}$ \\
    &&& k=5
    & $85.46_{0.07}$
    & $14.83_{0.06}$
    & $2.79_{0.12}$
    & $12.39_{0.02}$
    & $12.48_{0.10}$ \\
    \cmidrule{3-9}
    &&\multirow{4}{*}{Extractive}
    & k=0 & $85.75$ & $22.28$ & $14.26$ & $18.87$ & $14.60$ \\
    &&& k=1
    & $86.15_{0.35}$
    & $24.36_{1.77}$
    & $16.34_{1.77}$
    & $20.89_{1.70}$
    & $16.26_{1.44}$ \\
    &&& k=3
    & $86.15_{0.40}$
    & $24.01_{2.22}$
    & $16.07_{2.41}$
    & $20.66_{2.34}$
    & $15.99_{2.06}$ \\
    &&& k=5
    & $86.20_{0.39}$
    & $24.66_{2.06}$
    & $16.61_{2.09}$
    & $21.19_{2.02}$
    & $16.61_{1.87}$ \\
    \bottomrule
    \end{tabular}
    }
    \caption{Performance on argument generation tasks using \texttt{Flan-UL2}.}
    \label{tab:few-shot-AG-flan}
\end{table*}

\section{Case Study on Counter Speech Generation}
\label{sec:case}

Table \ref{tab:case} shows a case study of the data sample shown in Table \ref{tab:sample} for the proposed counter speech generation task.
The pipeline approach by extracting claims first tends to generate repetitive phrases, and does not attack all the claims stated in the supporting speech.
In contrast, the end-to-end approach is more concise and attacks the claims in the supporting speech.

\begin{table*}[ht]
	\centering
	\resizebox{\linewidth}{!}{
	    \setlength{\tabcolsep}{0mm}{
            \begin{tabular}{p{2.5cm}@{~} @{~}p{20cm}}
            \toprule
            Topic & Nationalism does more harm than good \\
            \midrule


            Pipeline 1 (Claim Detection)
            & 
Nationalism can actually have positive effects on a country and its citizens.
It can promote a sense of unity and pride in one's country, which can lead to increased civic engagement and a desire to work towards the betterment of the nation.
Additionally, nationalism can serve as a unifying force in times of crisis or conflict, bringing people together to overcome challenges and protect their country.
While extreme forms of nationalism can certainly lead to harm, it is important to recognize that a moderate level of nationalism can have beneficial effects. 
\textcolor{red}{While it is true that nationalism can be used as a tool for fascism, it is not inherently fascist.} National identity and community can be important for promoting unity and a sense of belonging among citizens. Additionally, a strong national identity can help to promote national pride and a sense of shared values and history. It is important to distinguish between healthy nationalism and dangerous nationalism, and to work towards promoting the former while rejecting the latter. While there are certainly negative aspects to nationalism, it is important to recognize that it can also have positive effects. Nationalism can foster a sense of unity and pride among a country's citizens, which can lead to increased social cohesion and a stronger sense of national identity. Additionally, nationalism can be a powerful force for promoting democracy and human rights, as citizens are more likely to demand these things from their government when they feel a strong sense of national identity and pride. Therefore, while nationalism should certainly be approached with caution, it is not necessarily a wholly negative political force. \textcolor{red}{While it is true that nationalism can sometimes lead to exclusionary policies, it is important to note that a sense of national identity and pride can also bring people together and promote unity.} Additionally, a strong national economy can provide resources and opportunities for all citizens, including refugees. It is not necessarily nationalism itself that harms disenfranchised people, but rather the actions and policies of those in power who use nationalism as a tool for exclusion and discrimination. Therefore, it is important to address and challenge these harmful actions rather than dismissing nationalism as inherently harmful. \textcolor{red}{While it is true that social media can create divisions within a society, it can also bring people together and foster a sense of community.} Social media platforms allow individuals to connect with others who share similar interests and beliefs, regardless of geographical location. This can lead to the formation of online communities that provide support, encouragement, and a sense of belonging. Additionally, social media can be used to raise awareness about important social issues and promote positive change. Therefore, it is important to recognize the potential benefits of social media in promoting social cohesion and unity. \textcolor{red}{While it is true that a strong sense of national pride can sometimes lead to corruption and threats to democracy,} it is important to note that not all instances of national pride result in negative consequences. In fact, a healthy sense of national pride can often lead to positive outcomes such as increased civic engagement, a stronger sense of community, and a greater willingness to work towards the common good. Additionally, it is important to recognize that corruption and threats to democracy can arise from a variety of factors, not just national pride. Therefore, it is not fair to solely blame national pride for these issues. Instead, we should focus on addressing the root causes of corruption and threats to democracy, while also promoting a healthy sense of national pride that is grounded in respect for diversity and democratic values. \textcolor{red}{While it is true that nationalism can sometimes lead to divisions within society,} it can also serve as a unifying force for a country. Nationalism can create a sense of pride and identity among citizens, which can lead to a stronger sense of community and a shared vision for the future. Additionally, nationalism can be a powerful tool for promoting social and economic progress, as it encourages people to work together towards a common goal. Therefore, while it is important to be mindful of the potential negative effects of nationalism, it is also important to recognize its potential benefits and to use it in a responsible and constructive way. While it is true that nationalism has been associated with negative outcomes such as war and discrimination, it is important to recognize that it has also played a positive role in shaping national identity and promoting unity among citizens. Nationalism has been a driving force behind important social and political movements, such as the fight for independence and civil rights. Additionally, a sense of national pride can inspire individuals to work towards the betterment of their country and its people. Therefore, it is unfair to dismiss nationalism as solely harmful without acknowledging its potential benefits.
            \\
            
            

            \midrule
            
            End-to-end
            & 
While I understand the concerns raised about nationalism, I believe that it is important to recognize the positive aspects of national identity. 
    \textcolor{red}{Nationalism can provide a sense of belonging and community for individuals who share a common culture, language, and history. }
    It can also promote a sense of pride in one's country and encourage individuals to work towards the betterment of their nation.  \textcolor{red}{Furthermore, it is important to note that nationalism does not necessarily have to be exclusionary.} A cosmopolitan nationalism, as mentioned, can celebrate diversity and promote inclusivity. It is possible to embrace one's national identity while also recognizing and respecting the identities of others.  \textcolor{red}{Additionally, the argument that nationalism leads to xenophobia and a reluctance to accept refugees is not necessarily true.} While there may be instances where nationalism is used to justify exclusionary policies, it is important to recognize that there are also many individuals and communities who embrace refugees and work towards creating a more inclusive society.  \textcolor{red}{Finally, the argument that nationalism leads to the rise of strongman leaders and threats to democracy is not necessarily a result of nationalism itself, but rather a result of individuals who use nationalism as a tool for their own political gain.} It is important to hold these individuals accountable for their actions and to work towards promoting a more inclusive and democratic society.  In conclusion, while there are certainly concerns and challenges associated with nationalism, it is important to recognize the positive aspects of national identity and to work towards promoting a more inclusive and diverse society.
            \\
            \bottomrule
            \end{tabular}
        }
    }
    \caption{A case study of the data sample (shown in Appendix \ref{sec:sample}) from the benchmark dataset for the counter speech generation task.}
\label{tab:case}
\end{table*}

\end{document}